%% file: root.tex
\documentclass[letterpaper, 10 pt, conference]{ieeeconf}
\IEEEoverridecommandlockouts
\usepackage{stfloats}
\usepackage{amsmath,amssymb}
\usepackage{booktabs}
\usepackage{multirow}
\usepackage{graphicx}
\usepackage{xcolor}
\usepackage{algorithm}
\usepackage{algpseudocode}
\usepackage{paralist}
\usepackage{url}
\usepackage{caption}
\usepackage{cite}
\usepackage{cuted}
\usepackage{tabularx}
\usepackage{booktabs}
\usepackage{array}
\usepackage[colorlinks=false,
            pdfborder={0 0 1},        
            citebordercolor={1 0 0},  
            linkbordercolor={1 1 1},  
            urlbordercolor={1 1 1}]{hyperref}
\usepackage{xcolor}   
\newcommand{\blueurl}[1]{\href{#1}{\textcolor{blue}{\nolinkurl{#1}}}}            

\makeatletter
\let\NAT@parse\undefined
\makeatother

\newcommand{\method}{\textsc{EmbodyHOI}}

\newcommand{\svmin}{\sigma_{\min}}

\title{\LARGE \bf EmbodyHOI: Embodiment-Aware Generation of Hand--Object Interactions for Humanoid Dexterous Manipulation}

\author{
  Chaitanya Chawla$^{*}$,\; Kartik Sharma$^{*}$,\; Ananya Bal,\; Laszlo A. Jeni,\; Guanya Shi%
  \thanks{$^{*}$Equal contribution.}\\
  Carnegie Mellon University \\
  United States\\
  \texttt{\{cchawla, kartiksh, abal, laszloaj, guanyas\}@andrew.cmu.edu} \\
}

\IEEEaftertitletext{\vspace{-2.0\baselineskip}}

\begin{document}
\maketitle
\thispagestyle{plain}
\pagestyle{plain}

\begin{strip}
\centering
\includegraphics[width=\textwidth]{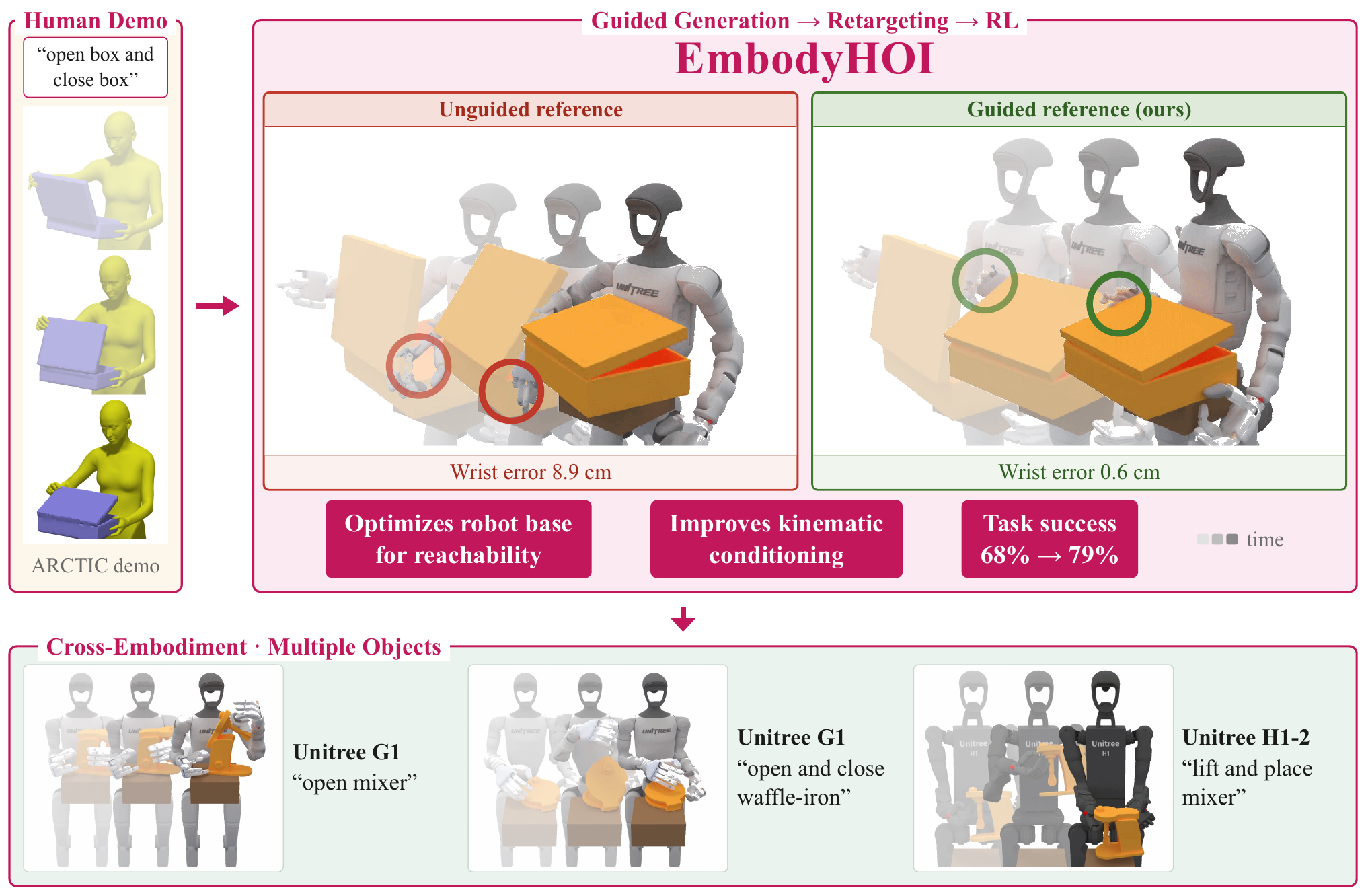}
\captionof{figure}{\method{} steers a human-space HOI trajectory generation method toward motions that are more reachable for specific humanoid embodiments. During diffusion sampling in the HOI generation, a differentiable IK solve on the target robot scores wrist trajectories, and its gradient is fed back into the sampler while the robot's base placement is optimized jointly. The generator is never retrained, so the same model and the same IK objective serve the Unitree G1 and H1-2.}
\label{fig:teaser}
\vspace{-8pt}
\end{strip}

\begin{abstract}
Humanoid dexterous manipulation needs reference motions that are both semantically rich and
kinematically feasible for the target robot.
Text-conditioned hand--object interaction (HOI) generators supply diverse, language-specified
demonstrations, but they synthesize motion in human (MANO) space.
The embodiment gap is therefore deferred to post-hoc retargeting, where trajectories deployed on the robot can fail to maintain contact with objects and also degrade policy learning.
Our key idea is that \textit{robot reachability} should constrain the interaction trajectories \emph{while they are being
generated}.
We present \method{}, which runs a differentiable inverse-kinematics (IK) solver for a specific humanoid embodiment inside the
sampling loop of a frozen flow-matching HOI trajectory generator. We feed back gradients from the embodiment-aware IK solver into the HOI sampler and jointly optimize the position of the humanoid with respect to the object.
Because guidance acts only at inference, one HOI trajectory generator trained on human hand poses transfers across humanoid embodiments without
retraining.
On ARCTIC trajectories with two contrasting humanoids (Unitree G1 and H1-2), our proposed guidance-based method cuts the retargeted
wrist error from 8.9\,cm to 0.6\,cm while leaving text-to-HOI generation quality intact, and
raises task success from 68\% to 79\% on 50 random G1 tasks
and from 62\% to 76\% on a four-task H1-2 suite.
Critically, cheaper alternatives for bridging the embodiment gap fail to beat EmbodyHOI. Post-hoc wrist refinement and
compute-matched best-of-$N$ selection both reach \emph{lower} IK energy than our guidance yet
train worse policies. For simulation results, see our website: \blueurl{https://embodyhoi.github.io/}
\end{abstract}


\section{Introduction}
\label{sec:intro}

Dexterous manipulation with multi-fingered hands remains difficult for humanoids that can reliably walk and loco-manipulate~\cite{he2025hover,liao2025beyondmimic,qiu2025humanoid,weng2025hdmi}.
A robot must coordinate reaching, finger contact, and whole-arm motion under embodiment-specific kinematic constraints to execute trajectories that actually manipulate objects.
A human trajectory when kinematically retargeted to a humanoid is often not executable, due to differences in finger morphology, joint limits, and wrist reachability.

We study how to \emph{generate} human interactions which can easily close the above gap. Given a language prompt and an
articulated object, we produce bimanual hand--object trajectories specific to a humanoid, which are easily executable by the robot.

A natural route to this goal of end-to-end policies is generate-then-track, but today's Hand-Object Interaction (HOI) generators have no notion of the robot that will execute their output. In this route, a generator proposes diverse task-conditioned reference motions and a physics-based RL policy learns to track them under contact and actuation limits~\cite{kalaria2025dreamcontrol}.

Hand-centric HOI generators are the natural source of dexterous references as they model finger-object contact, but they generate in MANO~\cite{romero2017mano} space with no term that couples the wrist to a robot's kinematic chain. This mismatch only surfaces during retargeting, after the finger and object motions are committed, and the tracking policy inherits wrist targets that the arm cannot reach. Previous works resolve this on either side of generation. Training the motion prior in robot space~\cite{harithas2026dreamcontrol} yields one generator per robot, which rules out cross-embodiment reuse. Whereas repairing a generated human motion~\cite{park2025demodiffusion,chen2024object} only makes delta modifications, and cannot always rectify retargeted wrist trajectories unreachable for the robot. Tracking systems such as DexMachina~\cite{mandi2025dexmachina} and ManipTrans~\cite{li2025maniptrans} sidestep the question by commanding a floating wrist with no arm.

We take a different approach: keep a frozen MANO HOI generator as a \emph{reusable} motion prior and make its samples embodiment-aware \emph{during generation} (Fig.~\ref{fig:teaser}).
We build on MAD-HOI~\cite{madhoi}, a text-conditioned HOI generator that fills object and hand tokens over a sequence of unmasking steps, integrating a flow-matching ODE for the unmasked tokens at each step. Inside every ODE step, we decode the current wrist trajectories, solve a differentiable IK problem on the target humanoid, and add a gradient of the IK residual (\emph{IK energy}) to the learned velocity field, in a manner of classifier guidance. A wrist correction made in one unmasking step conditions the finger and object tokens generated in later ones, so the interaction re-forms around the corrected wrist instead of being patched afterward.

A full humanoid raises another problem that floating hands and fixed-base arms do not, where the robot stands. The base placement decides whether both arms stay reachable over the whole interaction, and a fixed-base IK cannot differentiate between a feasible interaction performed from the wrong standing position from an infeasible one. We therefore solve for the base translation together with the arm joints, and re-solve it at every unmasking step so that the base follows the partially generated motion. Finally, we retarget the guided motions and train a PPO tracking policy with separate imitation rewards for arm and hand, as they differ in error scale and contact requirements. 

\emph{IK energy} is a diagnostic, not a quantity to minimize on its own.
Guidance costs $32\times$ a native sample (5\,s to 160\,s on an A100), paid once per reference, so we compare it with alternatives that lower the energy after generation at the same or lower cost.
Post-hoc IK refinement of a finished sample reaches the lowest energy of any method, and compute-matched best-of-$N$ selection also beats guidance on energy, yet both train worse policies (Table~\ref{tab:necessity}).
Refinement moves the wrists after the fingers and object are fixed, and selection returns a different interaction instead of correcting the requested one.
Reachability has to be imposed while the interaction is still being generated.

Our contributions are as follows.
\begin{compactitem}
\item \emph{Training-free IK guidance for a frozen HOI generator.}
A differentiable IK solve on the target humanoid runs inside every ODE step of the sampler, and its gradient steers the hand tokens.
One objective with one set of weights serves the Unitree G1 and H1-2, and it also lowers the IK energy of a second generator, Text2HOI~\cite{cha2024text2hoi}, without modification.
\item \emph{Base placement inside guidance.}
Solving for the base removes near-singular G1 arm configurations (2.6\% to 0.2\% of test samples) even when applied once to a finished sample, and re-solving it at every unmasking step accounts for 2.8 of the 5.7 success-rate points that guidance adds over post-hoc base optimization on the four-task G1 suite (Table~\ref{tab:ablation}).
\item \emph{Lower IK energy after generation does not give better references.}
Post-hoc refinement and compute-matched best-of-32 selection reach lower energy than guidance (0.35 and 1.21 vs.\ 1.81), yet their policies succeed less often (32.8\% and 74.0\% vs.\ 79.7\%, Table~\ref{tab:necessity}).
\item \emph{Downstream gains on two humanoids.}
Policies trained on guided references with decoupled hand and arm reward streams, raise success from 68\% to 79\% on 50 randomly drawn ARCTIC tasks on the G1 and from 62\% to 76\% on a four-task H1-2 suite.
We will release the guidance code, the generated references and the trained policies.
\end{compactitem}

\section{Related Work}
\label{sec:related}

\method{} draws on three lines of work, and none of them adapts the reference motion to a humanoid's arm chain while the interaction is being generated.

\textbf{Human-space HOI generation and test-time guidance.}

Text-to-motion models generate full-body SMPL~\cite{loper2015smpl} motion with pose-space diffusion~\cite{tevet2023mdm,zhang2022motiondiffuse} or masked-token generation~\cite{guo2024momask}.
Full-body HOI methods~\cite{xu2023interdiff,peng2023hoidiff} model contact only coarsely at the body surface.
Hand-centric methods parameterized by MANO~\cite{romero2017mano} model fine-grained finger--object contact on datasets such as ARCTIC~\cite{fan2023arctic}, GRAB~\cite{taheri2020grab}, and OakInk~\cite{yang2022oakink}, from grasp synthesis and interaction denoising~\cite{taheri2020grab,liu2024geneoh} to language-conditioned sequence generation~\cite{cha2024text2hoi,li2025latenthoi,zhang2026openhoi,christen2024diffh2o,madhoi}.
Test-time guidance steers such generators toward keyframes, trajectories, and contacts, but always toward targets specified in human space~\cite{karunratanakul2023gmd,xie2024omnicontrol}.
Our guidance has the same form but no human-space target: it scores the decoded wrist against a robot's kinematic chain, so its optimum moves when the robot or its base changes.

\textbf{Retargeting and RL from human demonstrations.}
Human-to-robot retargeting typically optimizes joint angles against keypoint and end-effector errors~\cite{kim2025pyroki,he2024omnih2o}, and it underpins teleoperation~\cite{he2024omnih2o,luo2025sonic} and robot-space datasets~\cite{harithas2026dreamcontrol}.
Dexterous-hand retargeting~\cite{qin2023anyteleop,pan2025spider,shaw2024leap} is a standard preprocessing step for RL from human demonstrations.
ManipTrans~\cite{li2025maniptrans} pretrains a hand-trajectory imitator and then fine-tunes a residual policy.
DexMachina~\cite{mandi2025dexmachina} uses retargeted joints and contacts as auxiliary rewards under a decaying virtual-object curriculum.
DexMan~\cite{hsieh2025dexman} extends this recipe to RGB video, and HERMES~\cite{yuan2025hermes} unifies heterogeneous motion sources for mobile bimanual manipulation.
In all of them the reference is fixed before the robot is considered.
ManipTrans and DexMachina command a floating 6-DoF wrist with no arm chain, and DexMan drives a G1 arm through IK but keeps the robot at the origin, so neither arm reachability nor base placement is ever optimized.
We reuse their object-side recipe (Sec.~\ref{sec:rl}) and make the reference itself the variable.

\textbf{Generative priors for policy learning.}
Example-guided RL~\cite{peng2018deepmimic,peng2021amp} extends to HOI imitation~\cite{wang2023physhoi,xu2025intermimic} and to tracking-as-reward execution~\cite{luo2024omnigrasp,kalaria2025dreamcontrol}.
The two works closest to ours resolve the embodiment mismatch after the motion exists.
Chen et al.~\cite{chen2024object} guide an RL finger controller with a generative prior that emits only wrists, fixed before the robot is considered, and DemoDiffusion~\cite{park2025demodiffusion} denoises a retargeted human trajectory with a robot diffusion policy that stays close to that trajectory and so inherits its reachability.
DreamControl-v2~\cite{harithas2026dreamcontrol} avoids retargeting by training in robot space, and fine-tuning a human-space generator on reachable data would likewise remove our inference cost; both need robot-feasible data for every embodiment, which a method like ours would have to produce, and yield one generator per robot.
However, none of them keeps one frozen human-space generator: we resolve reachability inside its sampling loop and handle base placement for a full humanoid arm chain.

\section{Method}
\label{sec:method}

\begin{figure*}[t]
    \centering
    \includegraphics[width=0.95\textwidth]{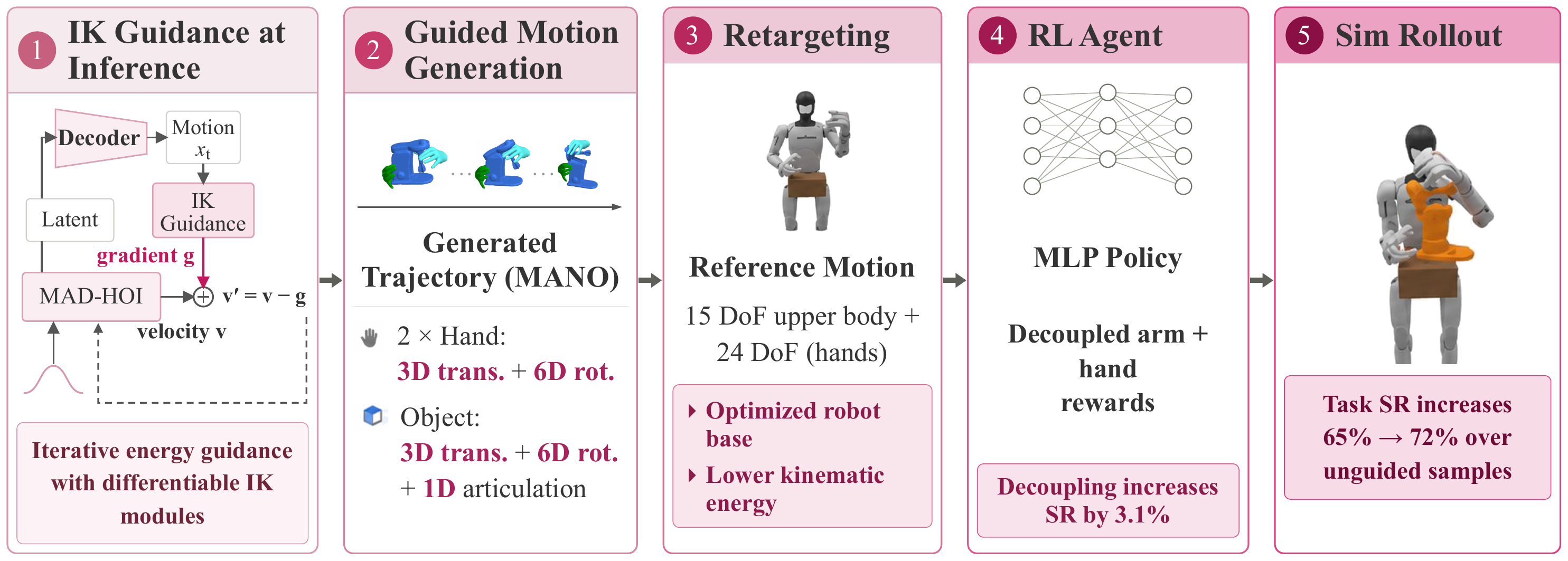}
    \caption{\method{} makes a frozen HOI generator embodiment-aware at inference. Inside every unmasking step of the masked autoregressive generator, a differentiable IK solve on the target humanoid scores the decoded wrist trajectories and places the robot's base; the gradient of the resulting IK energy is added to the flow-matching velocity field of the hand tokens, so the finger and object tokens generated in later steps form around reachable wrists. We then retarget the guided sample and train a PPO tracking policy on it with decoupled arm and hand rewards. Changing the robot changes only the IK module.}
    \label{fig:pipeline}
\end{figure*}

\method{} makes a \emph{frozen} HOI generator embodiment-aware at inference time (Fig.~\ref{fig:pipeline}).
The generator only needs to expose a decodable wrist pose and separate hand and object tokens (Sec.~\ref{sec:gen}).
A differentiable IK energy on the target humanoid steers its sampler and solves for the robot's base (Sec.~\ref{sec:ik}), and we retarget the guided motion and train a PPO tracking policy on it (Sec.~\ref{sec:rl}).

\subsection{Text-to-HOI Generation}
\label{sec:gen}
We adopt MAD-HOI~\cite{madhoi} as the generator and keep it frozen;
Sec~\ref{sec:cross} applies the same guidance to Text2HOI~\cite{cha2024text2hoi}.
Given a text prompt and an articulated object mesh, MAD-HOI produces an object trajectory $O=(\mathbf T^k_{\mathrm{obj}},\mathbf R^k_{\mathrm{obj}},\alpha^k_{\mathrm{obj}})$ (position, orientation, and articulation angle at frame $k$) and a MANO motion $H_h=(\boldsymbol\tau^k_h,\boldsymbol\phi^k_h,\boldsymbol\theta^k_h)$ for each hand $h\in\{L,R\}$: a wrist translation $\boldsymbol\tau^k_h\in\mathbb R^3$, a wrist orientation $\boldsymbol\phi^k_h\in SO(3)$, and 15 finger joint rotations $\boldsymbol\theta^k_h$, with rotations in 6D form.
Motion is represented in a decoupled latent space with separate object, left-hand, and right-hand tokens.
A masked autoregressive transformer generates these tokens over a sequence of unmasking steps, and a flow-matching head generates the tokens unmasked at each step.
The head is a learned velocity field $\mathbf v_\theta(\mathbf x,t,\mathbf z)$ that defines the ODE $\dot{\mathbf x}=\mathbf v_\theta(\mathbf x,t,\mathbf z)$ over the latent $\mathbf x$ of these tokens, with flow time $t\in[0,1]$ and a context $\mathbf z$ that the transformer computes from the already generated tokens and the text.
A decoder $\mathcal D$ maps latents back to motion.
Two properties of this generator make our method possible: the wrist pose can be decoded, differentiably, at any point of sampling, and hand and object tokens are separate, so a gradient can be applied to one and withheld from the other.
We use the ARCTIC dataset~\cite{fan2023arctic} (4.6k sequences, 11 articulated objects), whose bimanual articulated interactions are the hardest case for a humanoid arm chain.

\subsection{Embodiment-Aware IK Guidance}
\label{sec:ik}
Guidance acts on the wrists alone, through an IK energy on the target humanoid that also solves for the robot's base. Of the generated motion, only the wrist poses $\boldsymbol\tau^t_h\in\mathbb R^3$ and $\boldsymbol\phi^t_h\in SO(3)$ for $h\in\{L,R\}$ constrain the arm. 
Nothing ties these trajectories to the deployment robot, so per-frame IK retargeting is often ill-conditioned under large embodiment mismatch.
The manipulator Jacobian of arm $h$ with joint vector $\mathbf q^t$, $\mathbf J_h(\mathbf q^t)=\partial\,\mathrm{vec}(\mathbf T^t_{\mathrm{ee},h})/\partial\mathbf q^t\in\mathbb R^{6\times n}$, loses rank near singular configurations.
We monitor its minimum singular value $\svmin(\mathbf J_h)$ and condition number $\kappa(\mathbf J_h)$ as proximity measures, and bias \emph{generation} away from such configurations.

\textbf{Guidance adds an energy gradient to velocity field.}
Following classifier guidance~\cite{dhariwal2021diffusion}, adapted to a flow-matching velocity field, we subtract the gradient of an energy $E$ during sampling,
\begin{equation}
\dot{\mathbf x}=\mathbf v_\theta(\mathbf x,t,\mathbf z)-w\,\nabla_{\mathbf x}E(\mathbf x),
\label{eq:guidance}
\end{equation}
with guidance weight $w\ge0$. The energy is evaluated on the decoded wrists: $\mathcal D$ maps the current latent, together with the tokens already unmasked, to wrist trajectories $(\boldsymbol\tau,\boldsymbol\phi)$, which serve as task-space targets for an IK solve on the robot.

\textbf{Joint arm and base IK.}
We solve for the base because a fixed base makes reachability a property of the
trajectory alone: a feasible interaction performed from the wrong standing position is then
indistinguishable from an infeasible one, and fixed-base IK rejects both.

At each timestep we optimize $\mathbf u^t=[\mathbf q^t;\mathbf b^t]$, arm joints and a base translation $\mathbf b^t\in\mathbb R^3$ (base orientation fixed). With world-frame wrist target $\mathbf T^t_{w,h}$ built from $(\boldsymbol\tau^t_h,\boldsymbol\phi^t_h)$, the base transform $\mathbf T^b$ built from $\mathbf b^t$, and the forward kinematics $\mathrm{FK}_h(\mathbf q^t)$ of wrist $h$ in the base frame, the pose residual of arm $h$ is
\begin{equation}
\mathbf e^t_h=\log\big((\mathbf T^t_{w,h})^{-1}\mathbf T^b\,\mathrm{FK}_h(\mathbf q^t)\big)
\in\mathbb R^6
\label{eq:residual}
\end{equation}
weighted by $w_{\mathrm{pos}}$ on its translational part and $w_{\mathrm{ori}}$ on its rotational part.
The regularizers complete the residual and fall into the two groups that Table~\ref{tab:ablation} ablates.
The \emph{base-placement group} acts on $\mathbf b^t$: a base prior that penalizes displacement from the current preferred placement $\mathbf b_0$ ($w_{\mathrm{root}}$), and a clearance term that keeps the base outside a fixed horizontal radius around the object ($w_{\mathrm{clear}}$).
The \emph{arm-conditioning group} acts on $\mathbf q^t$: a rest-pose prior ($\lambda_{\mathrm{rest}}$), a joint-limit penalty ($w_{\mathrm{joint}}$), a penalty on deviation from the previous solve's configuration, which also warm-starts the solve ($w_{\mathrm{smooth}}$), and a null-space manipulability step ($\lambda_{\mathrm{manip}}$).
The null-space step runs after each solve.
It takes a few gradient steps on $1/\svmin(\mathbf J_h)$, where $\mathbf J_h(\mathbf q)\in\mathbb R^{6\times n}$ is the geometric Jacobian of wrist $h$ and $\svmin$ its minimum singular value, projected onto the null space of both wrist tasks; it thus improves manipulability without moving either wrist, and the refined configuration seeds the next solve.
Because $\mathbf J_h$ loses rank at singular configurations, $\svmin$ and the condition number $\kappa(\mathbf J_h)$ are also the conditioning measures we report in Sec.~\ref{sec:exp}.

Stacking residuals into $\mathbf r(\mathbf u)$ and solving with Levenberg--Marquardt (10 iterations) gives $\mathbf u^{t*}$, and the IK energy is the mean squared residual over the horizon of $T$ timesteps,
\begin{equation}
E(\mathbf x)=\tfrac1T\textstyle\sum_{t=1}^T\|\mathbf r(\mathbf u^{t*})\|_2^2 .
\label{eq:energy}
\end{equation}
All weights are set once and shared by both embodiments, both generators, and all tasks (Table~\ref{tab:ik_guidance}).

Although the solve optimizes a base translation per timestep, the base is fixed during execution, so we average the per-timestep estimates into a single execution pose $\mathbf b_0\leftarrow\mathrm{mean}_t(\mathbf b^{t*})$, which becomes the root prior for the next unmasking iteration (last line of Alg.~\ref{alg:outer}).
The prior starts from one hand-set placement per robot and settles as the sequence fills in: its median change per step falls from about 1.3\,cm in the first unmasking steps to 0.06\,cm in the last.
Because the base is solved per robot, the same generator can serve robots of different proportions: a 45\,cm G1 arm and a 68.5\,cm H1-2 arm must stand in different places for the same interaction (Sec.~\ref{sec:cross}).

\textbf{Gradients through the solver.}
We differentiate implicitly through the LM solve in JAX, so $\partial E/\partial(\boldsymbol\tau,\boldsymbol\phi)$ is defined without unrolling the iterations, and a JAX--PyTorch bridge propagates it through $\mathcal D$ to the latent $\mathbf x$.
Before the update, we zero the gradient on object tokens ($\mathbf m_{\mathrm{hand}}$ in Alg.~\ref{alg:outer}), so guidance never acts on object tokens directly and the generator's decoupled object--hand structure is preserved.
Object and finger tokens generated in later unmasking steps still condition on the guided hand tokens, which is how the interaction re-forms around the corrected wrists.

\begin{algorithm}[t]
\caption{One outer unmasking step with IK guidance}
\label{alg:outer}
\small
\begin{algorithmic}[1]
\Require partial latents $\mathbf L$, outer step $s$, IK module $\mathcal G$, weight $w$, base prior $\mathbf b_0$
\State sample mask $\mathcal M$ (cosine schedule); build $\mathbf L_{\mathrm{mask}}$
\State $\mathbf z_{\mathcal M}\gets\mathrm{MAD\text{-}HOI}(\mathbf L_{\mathrm{mask}},\mathbf c_{\mathrm{text}})$; $\mathbf x\sim\mathcal N(\mathbf 0,\mathbf I)$
\For{flow time $t=0\to1$ (RK4, 20 steps)}
  \State $\mathbf v\gets\mathrm{MLP}(\mathbf x,t,\mathbf z_{\mathcal M})$;\quad $(\boldsymbol\tau,\boldsymbol\phi)\gets\mathcal D(\mathbf x;\mathbf L_{\mathrm{mask}})$
  \State $E\gets\mathcal G(\boldsymbol\tau,\boldsymbol\phi;\mathbf b_0)$ \Comment{joint arm+base LM solve, Eq.~\eqref{eq:energy}}
  \State $\mathbf x\gets\mathrm{ODEStep}\big(\mathbf x,\ \mathbf v-w\,\nabla_{\mathbf x}E\odot\mathbf m_{\mathrm{hand}},\ t\big)$
\EndFor
\State $\mathbf L[\mathcal M]\gets\mathbf x$
\If{$s\ge s_{\mathrm{start}}$} decode $\mathbf L$; $\mathbf b_0\gets\mathrm{mean}_t(\mathbf b^{t*})$ \EndIf
\end{algorithmic}
\end{algorithm}

\textbf{Guidance must act during generation.}
We inject guidance into the inner ODE of every unmasking step (Alg.~\ref{alg:outer}) and refresh the base prior after each.
Guidance therefore edits the wrist while finger and object tokens are still being generated, so the interaction adapts to the new wrist; it changes the sampling distribution instead of ranking its draws.
Section~\ref{sec:necessity} tests two alternatives that lower the same energy after generation, post-hoc refinement of a finished sample and best-of-$N$ selection: both reach lower $E$ than guidance, and both train worse policies.

\subsection{Retargeting and Policy Learning}
\label{sec:rl}
\textbf{Retargeting turns the guided sample into a robot reference.}
Unless stated otherwise, the target is a Unitree G1 with two XHands: 15 controllable upper-body DoFs (torso and both arms), fixed lower body, and 12 DoFs per hand.
PyRoki~\cite{kim2025pyroki} solves the guided wrist trajectories to arm joint trajectories from the optimized base; dex-retargeting~\cite{qin2023anyteleop} maps MANO fingers to XHand joints; and contact retargeting following DexMachina~\cite{mandi2025dexmachina} yields the reference contacts $\hat{\mathbf c}_t$.
Generated object motion supplies the reference object pose and articulation used in observations, rewards, and terminations.

\textbf{The policy tracks the reference with PPO.}
We train an MLP policy with PPO~\cite{schulman2017proximal} in Genesis.
It observes short histories of joint positions, velocities, and actions; the episode phase; sparse future reference joint offsets $\hat{\mathbf q}^{\mathrm{robot}}_{t+\Delta}-\mathbf q^{\mathrm{robot}}_t$ for $\Delta\in\{1,2,4,8,16\}$; and the object's part poses, articulation, and future tracking errors against the generated trajectory.
Actions are 39-D: increments on the previous joint target for the 15 upper-body DoFs, which keep large arm motions smooth, and absolute targets for the 24 finger DoFs, whose joint ranges are small.
We never actuate the object, so unlike DexMachina we use no virtual-object-controller curriculum.

\textbf{Arm and hand get separate imitation rewards and terminations.}
Coarse free-space reaching and contact-rich finger control differ in error scale, so a single imitation term over all 39 DoFs biases optimization toward whichever subsystem dominates the error.
We therefore split imitation into an upper-body reward $r_{\mathrm{imi,upper}}$ over 15 DoFs and a dexterous reward $r_{\mathrm{imi,dex}}$ over 24 DoFs, each a joint-tracking error against the retargeted reference with its own weight and temperature.
The total reward adds an object-tracking term $r_{\mathrm{task}}$ on position, orientation, and articulation, a contact term that matches the retargeted contact locations, and standard penalties on action magnitude, action rate, joint velocity, joint acceleration, and torque.
Terminations are decoupled in the same way, on the half-squared joint error $\tfrac12\Delta q^2$: the upper-body error terminates on its \emph{maximum} over joints, because a large arm failure ends the rollout, and the finger error on its \emph{mean}, because an isolated finger error is not catastrophic.
Episodes also end if the object falls below the table or its half-squared articulation error exceeds $\epsilon_a$.

\subsection{Implementation Details}
\label{sec:impl}
The guided sampler runs 18 unmasking steps.
Inside each, it integrates the flow ODE with RK4 in 20 steps and applies guidance with a constant weight $w{=}0.5$ at every ODE step, from the first unmasking step on ($s_{\mathrm{start}}{=}0$).
Each IK solve runs LM for 10 iterations with implicit differentiation.
Appendix~\ref{app:ik} lists the objective weights (Table~\ref{tab:ik_guidance}) and solver settings (Table~\ref{tab:ik_hyperparams}), which are identical for both embodiments, both generators, and every task.
Each policy trains for about 4\,h on one RTX~4090.
Appendix~\ref{app:rl} gives the full RL specification: observations (Table~\ref{tab:observations}), reward terms (Table~\ref{tab:reward_terms}), terminations (Table~\ref{tab:termination_conditions}), and all PPO, reward, penalty, threshold, and action-parameterization values (Table~\ref{tab:hyperparams}).
The same RL settings are used for every condition in Tables~\ref{tab:necessity} and~\ref{tab:ablation}; only the reference differs.

\section{Experiments}
\label{sec:exp}

We answer four questions.
(Q1)~Does guidance \emph{during} sampling earn its cost relative to post-hoc IK refinement and compute-matched best-of-$N$ selection? (Sec.~\ref{sec:necessity})
(Q2)~How does guidance change kinematic conditioning and generation quality? (Sec.~\ref{sec:quality})
(Q3)~How much of the downstream gain comes from each component? (Sec.~\ref{sec:downstream})
(Q4)~Does the same objective transfer across embodiments and generators? (Sec.~\ref{sec:cross})

\textbf{Protocol.}
Downstream results use two protocols on the G1 with XHands: a fixed four-task suite (box open, waffle-iron lift, mixer open+close, microwave lift+place; Tables~\ref{tab:necessity} and~\ref{tab:ablation}) and 50 randomly sampled tasks with no pre-selection.
Of the latter, the 10 whose unguided reference falls below $\svmin{<}0.1$, and consequently carries high IK energy, form a hard subset (Sec.~\ref{sec:downstream}).
Retargeting, rewards, and PPO follow Sec.~\ref{sec:rl} in every condition; only the reference differs.
Q1 uses 51 seed-matched test prompts that include the four-task suite, Q2 400 ARCTIC test prompts, Q3 the four-task suite and the 50 random tasks, and Q4 the prompt sets stated in Sec.~\ref{sec:cross}.
We recompute every IK metric for every condition with one standalone instance of the IK solver of Sec.~\ref{sec:ik} (optimized base, two base-prior updates from the same initial prior), so the rows of a table are directly comparable.

Success requires the object's position, orientation, and articulation to stay within 2\,cm, 40$^\circ$, and 10$^\circ$ of the reference at every step of the episode.
Success rates (SR) are means over 5 training seeds, each evaluated on 10 episodes of $B$ parallel environments with randomized object initialization per task ($10B$ rollouts per seed and task), with the standard error over seeds in Tables~\ref{tab:necessity} and~\ref{tab:ablation}.
Kinematic metrics are the IK energy $E$ (Eq.~\ref{eq:energy}) and the worst-case Jacobian conditioning over the sequence (minimum singular value $\svmin$, maximum condition number) on the target robot's upper-body chain.

Generation quality uses the standard text-to-motion metrics (FID, KID, diversity, top-3 retrieval accuracy, matching score) and geometric metrics (contact ratio, penetration depth, jitter).

\subsection{Is Guidance During Generation Necessary? (Q1)}
\label{sec:necessity}

\begin{table}[t]
\centering
\caption{Guidance during sampling trains the best policies, although two alternatives reach lower IK energy. Medians over 51 seed-matched prompts. Cost: sampling time per reference in native samples ($t_1{=}5$\,s); $\Delta_{\mathrm{wrist}}$: wrist displacement (cm) from the seed-matched unguided sample; Pen.: penetration loss; SR: success rate on the four-task suite, mean $\pm$ s.e.m.\ over 5 seeds.}
\label{tab:necessity}
\footnotesize
\setlength{\tabcolsep}{1.8pt}
\begin{tabular}{lccccc}
\toprule
Method & Cost & $E\downarrow$ & $\Delta_{\mathrm{wrist}}$ (cm) & Pen.$\downarrow$ & SR\,\%$\uparrow$ \\
\midrule
Unguided, fixed base   & $t_1$    & 2.30 & --    & 0.52 & 68.1 $\pm$ 0.3 \\
+ base optimization     & $t_1$    & 1.89 & 0     & 0.52 & 73.3 $\pm$ 0.5 \\
+ IK guidance (ours)    & $32\,t_1$ & 1.81 & \textbf{0.3} & 0.53 & \textbf{79.7 $\pm$ 0.4} \\
\midrule
Post-hoc IK             & $t_1$+IK & \textbf{0.35} & 32.6 & 0.59 & 32.8 $\pm$ 0.3 \\
Best-of-8 by $E$        & $8\,t_1$  & 1.36 & 13.2 & 0.52 & 73.1 $\pm$ 0.6 \\
Best-of-32 by $E$       & $32\,t_1$ & 1.21 & 13.7 & 0.51 & 74.0 $\pm$ 0.6 \\
\bottomrule
\end{tabular}
\end{table}

Guidance costs $32\times$ a native sample, so each alternative in Table~\ref{tab:necessity} receives the same budget, the same 51 prompts, and the same IK solver and base optimizer.
\textbf{Base-placement} optimization alone keeps the sample and re-solves only the base. This cost is paid once per reference and amortized over $\approx$4\,h of policy training it supports, so the wall-clock comparison is a worst case rather than a deployment constraint.
\textbf{Post-hoc IK} refinement optimizes the wrist channels of a finished unguided sample against the same energy (Adam, 100 steps).
\textbf{Best-of-$N$} draws $N$ unguided samples with fixed seeds and keeps the one with the lowest terminal $E$ under the same optimized base.

\begin{figure}[t]
    \centering
    \includegraphics[width=0.92\columnwidth]{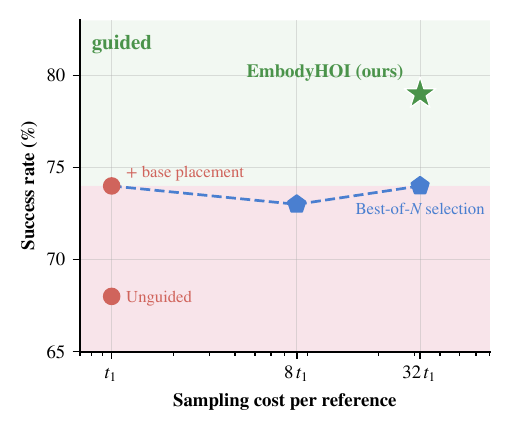}
    \captionsetup{skip=-6pt}
    \caption{Only guidance improves on base optimization. Success rate on the four-task suite against sampling cost per reference (Table~\ref{tab:necessity}). Base optimization lifts the unguided baseline at no sampling cost; best-of-$N$ selection does not measurably improve on it even at 32 times the cost ($0.7\pm0.8$ points).}
    \label{fig:compute}
\end{figure}

\textbf{Base placement is the cheapest large gain.}
Solving for the base with a single unguided sample reduces the median energy from 2.30 to 1.89 and raises the success rate from 68.1 to 73.3 at no sampling cost (Fig.~\ref{fig:compute}).
Guidance adds a smaller, consistent correction on top: it lowers the energy of the seed-matched sample on 40 of the 51 prompts (78\%; median 1.81) and raises the success rate to 79.7.

\textbf{IK energy is not the quantity to optimize.}
Energy is diagnostic only at fixed interaction content: it is meaningful for a
seed-matched guided-vs-unguided pair, and meaningless across independently drawn samples.
Post-hoc refinement reaches the lowest energy of any method ($E{=}0.35$) but moves the wrists by 33\,cm after finger and object motion are fixed, raising penetration from 0.52 to 0.59 and collapsing the success rate to 32.8.
Selection lowers the energy further than guidance at every budget: 1.36 at $N{=}8$ and 1.21 at the compute-matched $N{=}32$, and wins on 88--94\% of prompts --- 88\% on the easiest third by unguided $E$, 94\% on
the hardest --- so its advantage does not shrink where reachability is worst. (see Section~\ref{app:downstream}).

The reason is variance.
Energy varies across samples of one prompt (s.d.\ 0.36) far more than guidance shifts it (median per-prompt shift 0.04), so drawing more samples finds a low-energy trajectory more easily than correcting one sample does.
Yet the policies trained from selected references succeed less often than from guided ones (73.1 at $N{=}8$, 74.0 at $N{=}32$, vs.\ 79.7), a gap of $5.7\pm0.7$ points at matched compute.
The energy measures wrist reachability only; it is blind to whether fingers and object still form a coherent interaction around the wrists it prefers.

\textbf{Guidance corrects, selection re-draws.}
The selected best-of-$N$ sample lies as far from the seed-matched one as any other sample (13\,cm), whereas guidance moves the wrists of the seed-matched sample by 0.3\,cm and leaves penetration and contact unchanged.
Only guidance edits the sample already in hand, which is what matters when that sample is one a user chose.
Base-placement optimization is therefore the baseline to report alongside guidance: it is free, selection is a no-op on top of it at either budget ($-0.2\pm0.8$ at $N{=}8$, $0.7\pm0.8$ at $N{=}32$), and raising $N$ from 8 to 32 does not help either ($0.9\pm0.8$).

\subsection{Kinematic Conditioning and Generation Quality (Q2)}
\label{sec:quality}

\begin{figure*}[t]
    \centering
    \includegraphics[width=0.7\textwidth]{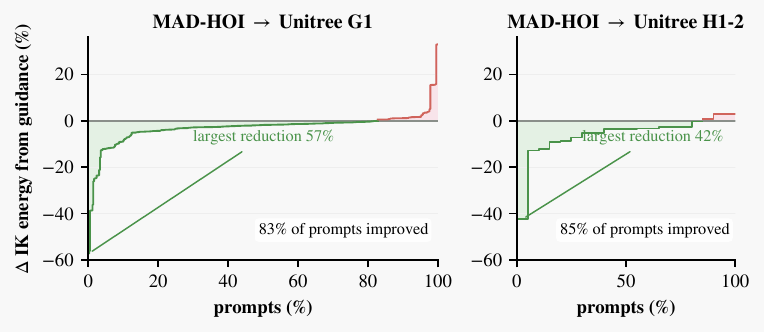}
    \caption{Change in IK energy that guidance applies to each seed-matched sample. Guidance
  makes 83--97\% of samples more reachable depending on the robot--generator pair; the remaining increases are small (median 1\%), with two
  outliers on the G1 where sampling propagated the correction into a different trajectory (Sec.~\ref{sec:quality}).}
    \label{fig:singularity}
    \vspace{-8pt}
\end{figure*}

\begin{figure}[t]
    \centering
    \includegraphics[width=\columnwidth]{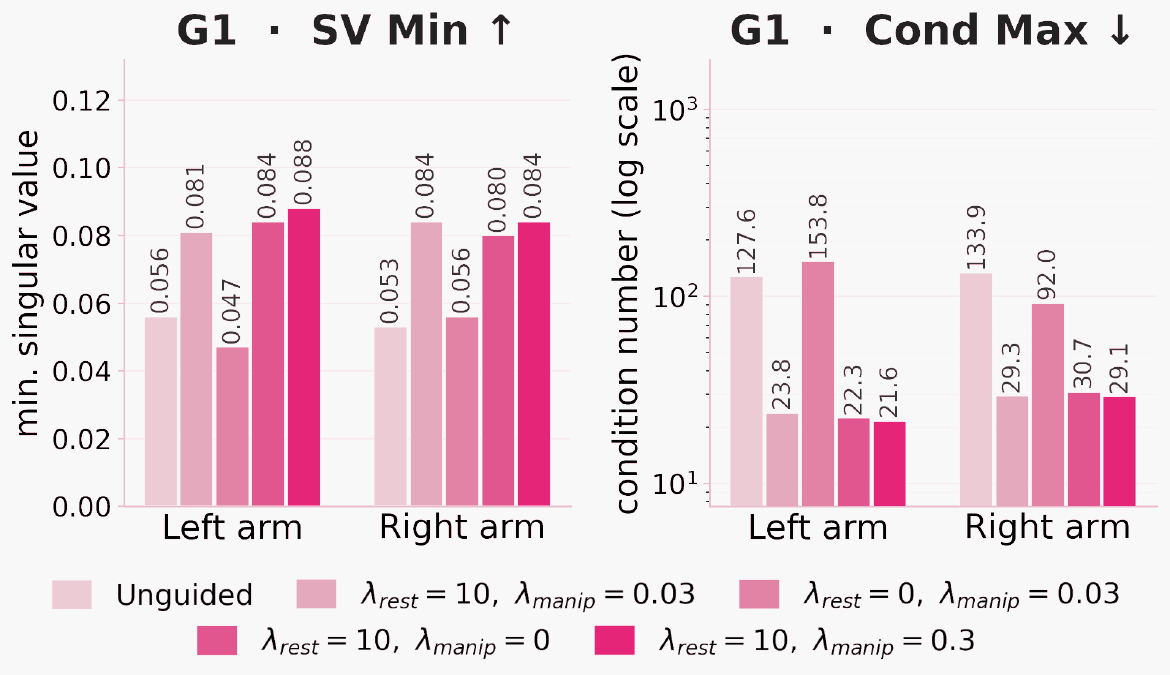}
    \caption{Base placement, not the IK weights, removes near-singular configurations. Worst-case conditioning of the G1 per arm over the ARCTIC prompts: minimum Jacobian singular value ($\uparrow$) and maximum condition number ($\downarrow$, log scale). Unguided samples are solved at the fixed base; guided samples are solved at the optimized base under four IK-weight settings, whose distributions overlap.}
    \label{fig:singgrid}
\end{figure}

\begin{table*}[t]
\centering
\caption{Guidance removes most of the retargeting error with little change in generation quality. Left: generation quality of unguided and IK-guided MAD-HOI on the \emph{same} 400 ARCTIC test prompts with one evaluator. Right: wrist tracking error of the retargeted G1 after PyRoki IK on the randomly sampled tasks.}
\label{tab:quality}
\small
\setlength{\tabcolsep}{4pt}
\begin{tabular}{lcccccccc|cc}
\toprule
 & FID$\downarrow$ & KID$\downarrow$ & Div.$\uparrow$ & Top-3$\uparrow$ & Match$\downarrow$ & Contact\,\%$\uparrow$ & Pen.\,$<$1\,cm\,\%$\downarrow$ & Jitter$\downarrow$ & Wrist err.\ (cm)$\downarrow$ & Wrist err.\ ($^\circ$)$\downarrow$ \\
\midrule
Unguided  & 0.667 & \textbf{0.001} & 19.77 & 0.852 & 3.99 & \textbf{83.5} & 12.2 & 0.287 & 8.9 & 20.2 \\
IK-guided & \textbf{0.545} & 0.003 & \textbf{19.85} & \textbf{0.872} & \textbf{3.97} & 83.1 & \textbf{11.5} & 0.293 & \textbf{0.6} & \textbf{4.0} \\
\bottomrule
\end{tabular}
\end{table*}

\textbf{Conditioning.}
  Base placement and guidance remove different failure modes.
  On 400 ARCTIC test prompts, optimizing the base alone cuts the share of near-singular unguided samples ($\svmin<0.03$) from 2.6\% to
  0.2\% and the median IK energy from 2.59 to 2.19 (Fig.~\ref{fig:singgrid}); after that, $\svmin$ ($\approx$0.086) and the condition
  number ($\approx$26) are as good as the chain allows, and guidance leaves them there.
  Guidance instead removes the residual error that survives retargeting: it lowers the energy on 83\% of prompts, by 2.9\% on average and
  by 8\% on the hardest quartile (Fig.~\ref{fig:singularity}), and the G1's wrist tracking error after retargeting falls from 8.9\,cm and
  20$^\circ$ to 0.6\,cm and 4$^\circ$ (Table~\ref{tab:quality}).
  On the remaining 17\% of prompts the energy rises by 1.2\% at the median with the wrists within 0.5\,cm of the unguided sample; on 2 of
  the 51 seed-matched prompts of Sec.~\ref{sec:necessity} the guidance step propagated through later unmasking rounds into a different trajectory.
  No sample is filtered, so the success rates in Secs.~\ref{sec:necessity} and~\ref{sec:downstream} include these cases. We do not measure a per-prompt relation between energy and success.

\textbf{Generation quality.}
Guidance does not degrade the prior (Table~\ref{tab:quality}).
On the identical 400-prompt subset, guided samples have lower FID than unguided (0.545 vs.\ 0.667), better top-3 retrieval (0.872 vs.\ 0.852), unchanged contact ratio (83.1 vs.\ 83.5\%) and diversity (19.85 vs.\ 19.77), lower penetration (11.5 vs.\ 12.2\%), and marginally higher KID (0.003 vs.\ 0.001).
Guidance necessarily integrates the flow ODE itself (20 RK4 steps, Alg.~\ref{alg:outer}), so this compares our sampler against MAD-HOI's released one rather than ablating the IK term alone, and the 0.12 FID gain belongs to the guided sampler as a whole.
The IK objective reshapes wrist trajectories toward reachable configurations without collapsing the semantic content of the interaction (see supplementary video).

\textbf{Sensitivity to weights.}
The weights are set once before the sweep and frozen across all tasks and both robots (Sec.~\ref{sec:cross}).
Zeroing the rest-pose prior, zeroing or tripling the manipulability term, or scaling the root prior by 0.3 on the same 400 prompts moves the median IK energy by at most 0.03 and $\svmin$ by less than 1\% (Fig.~\ref{fig:singgrid}), and every setting degrades FID (0.615--0.869 vs.\ 0.545) and top-3 retrieval (0.830--0.869 vs.\ 0.872). The manipulability term is the most sensitive of the four: tripling
$\lambda_{\mathrm{manip}}$ accounts for the
0.869 end of the FID range.
We did not retrain policies under perturbed weights, so downstream sensitivity is bounded only by Table~\ref{tab:ablation}, where dropping a whole weight group costs 2.9 to 4.0 points.

\subsection{Downstream Policy Learning and Ablations (Q3)}
\label{sec:downstream}

\begin{table}[t]
\centering
\caption{Guidance, in-loop base placement, and the decoupled reward each raise success on the G1 four-task suite (box open, waffle-iron lift, mixer open+close, microwave lift+place); mean $\pm$ s.e.m.\ over 5 seeds, 10 episodes per task. Rows A--D form a ladder: A$\to$B changes the reference source, B$\to$C changes the RL recipe (DexMachina-style~\cite{mandi2025dexmachina} vs.\ ours) and adds post-hoc base optimization, and C$\to$D adds IK guidance with in-loop base placement. The lower half removes one component at a time from D. Row D and the IK-guidance row of Table~\ref{tab:necessity} are independent training runs of the same configuration (79.3 vs.\ 79.7, within seed error).}
\label{tab:ablation}
\footnotesize
\setlength{\tabcolsep}{3pt}
\begin{tabular}{clc}
\toprule
 & Reference $\rightarrow$ retarget / RL & Avg.\ SR\,\%$\uparrow$ \\
\midrule
A & ARCTIC GT, fixed base $\rightarrow$ DexMachina-style & 66.8 $\pm$ 0.3 \\
B & Unguided, fixed base $\rightarrow$ DexMachina-style & 68.2 $\pm$ 0.9 \\
C & Unguided, optimized base $\rightarrow$ ours & 73.6 $\pm$ 0.9 \\
D & Guided (both groups) $\rightarrow$ ours (full) & \textbf{79.3 $\pm$ 0.6} \\
\midrule
 & D $-$ arm-conditioning group        & 76.4 $\pm$ 0.3 \\
 & D $-$ base-placement group          & 75.3 $\pm$ 0.3 \\
 & D, shared imitation reward          & 74.1 $\pm$ 0.6 \\
 & D, shared reward, no contact retarg.\ & 73.0 $\pm$ 0.4 \\
\bottomrule
\end{tabular}
\end{table}

\textbf{Randomly sampled tasks.}
Guided references raise the average success rate on 50 randomly sampled tasks from 68\% to 79\%, and on the 10 worst-conditioned of them --- selected by $sv_{min}$ of the
\textbf{unguided} reference before any policy is trained --- from 57.9\% to 82.1\%.
The 50 tasks are drawn with no selection across all 11 ARCTIC objects, and we train policies from a guided and an unguided reference of each (5 seeds).
The gain concentrates where reachability is poor: most randomly drawn prompts are already reachable, and the aggregate gain comes from the poorly conditioned minority (Sec.~\ref{sec:necessity}).

\textbf{Controlled comparison and ablation.}
Two changes move downstream success on the four-task suite, and they are comparable in size (Table~\ref{tab:ablation}).
A$\to$B holds reachability fixed and varies only the reference source: swapping the ARCTIC demonstration for an unguided generated sample changes the success rate by 1.4 points, so generation supplies diversity but not reachability.
B$\to$C then introduces our RL recipe together with base placement and adds 5.4 points, and C$\to$D adds IK guidance on top of it for a further 5.7.
Neither half alone accounts for the method's gain.
The ``DexMachina-style'' rows re-implement their object-side recipe on our 15-DoF arm chain; they are not a run of their released floating-wrist system.
Because that system commands a free-floating wrist with no arm chain, it never faces the reachability problem and is not a like-for-like baseline: its 86.1\% on the same tasks is an upper bound obtained by assuming away the constraint we study, and the gap to our 79.3\% is that constraint's cost rather than a deficiency of the RL recipe.

The two weight groups of the IK objective contribute comparably to the 5.7 points that C$\to$D adds.
Zeroing one group at a time, dropping base placement costs 4.0 points (79.3 to 75.3) and dropping the four arm-conditioning terms costs 2.9 (79.3 to 76.4); read the other way, base placement alone adds 2.8 over row C and arm conditioning alone 1.7, so the two together achieve more than the sum of their parts.
The comparison against row C is the one that matters for the design: row C applies the same base optimization once, after generation, so re-solving the base inside the sampling loop rather than after it (Sec.~\ref{sec:ik}) accounts for 2.8 of those points.
Base placement is therefore not preprocessing that any pipeline could bolt on, but a term in the guidance objective whose value comes from being inside the loop.
A shared imitation reward costs 5.2 points, and dropping contact retargeting only costs 1.1 more.

\begin{figure}[t]
    \centering
    \includegraphics[width=\columnwidth]{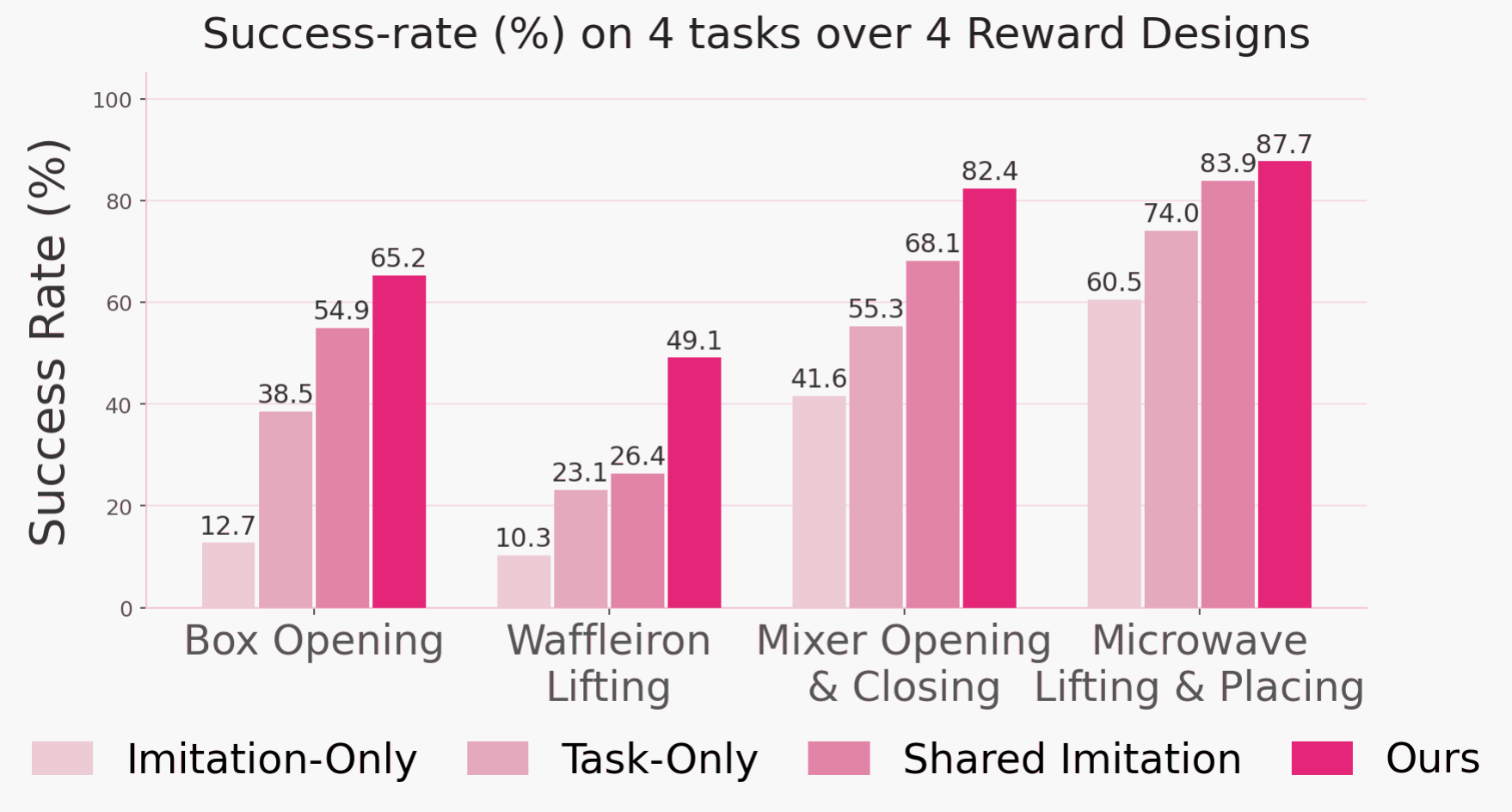}
    \caption{Decoupled arm and hand imitation rewards give the highest success rate on every task of the G1 four-task suite. Per-task success rate under four reward configurations: imitation only, object tracking only, one imitation term over all 39 DoFs, and our decoupled rewards.}
    \label{fig:app_reward}
\end{figure}

\textbf{Reward design.}
Averaged over the suite, imitation alone reaches 44.5\%, the object-tracking term alone 58.2\%, a single imitation term over all 39 DoFs 74.1\%, and our decoupled arm and hand streams 79.3\%.
Decoupling wins on every task, with the largest margins where coarse reaching and finger contact differ most in error scale.
Figure~\ref{fig:app_reward} gives the per-task breakdown behind the aggregate reward-design numbers.

\textbf{Feasibility beyond kinematics.}
Guidance does not reason about dynamics, but reachable references are also closer to executable ones: across the prompts of Sec.~IV-A, guided references have a lower median peak-velocity
ratio against the G1's joint limits than their seed-matched unguided counterparts (0.20 vs.\
0.24).

\subsection{Transfer Across Embodiments and Generators (Q4)}
\label{sec:cross}

\begin{figure}[t]
    \centering
    \includegraphics[width=0.78\columnwidth]{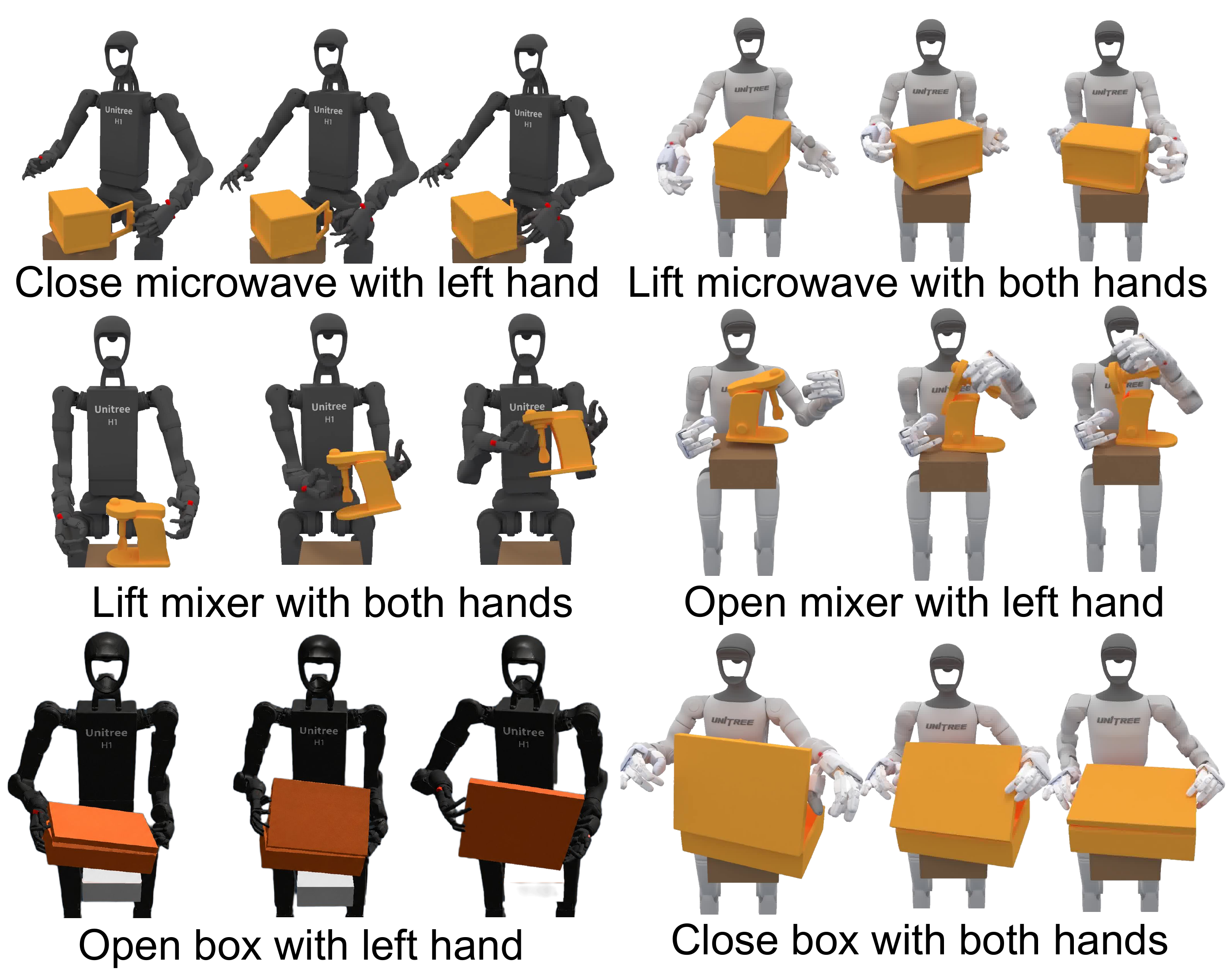}
    \caption{The same generator and IK objective place two humanoids differently: from the same initial placement, guidance moves the H1-2 base (left; 178\,cm, 68.5\,cm arms) away from the object to avoid over-extension and the G1 base (right; 132\,cm, 45\,cm arms) closer and higher to stay reachable.}
    \label{fig:cross}
\end{figure}

As guidance acts at inference on a frozen generator, shifting to a new robot only needs change in the IK module.
On the Unitree H1-2, whose arms are 68.5\,cm long compared to G1's 45\,cm, guidance lowers the IK energy of the seed-matched sample on 46 of 50 prompts (median $2.06\to1.95$, mean $-6\%$) while moving wrists by $0.3$\,cm.
Both conditions stay far from singularity ($\svmin\approx0.13$) because H1-2's longer arms are closer to human proportions.

Notably, the base moves in the opposite direction to G1's
(Fig.~\ref{fig:cross}): the longer arms over-extend and must stand back, where
the G1's must close in. No term in the objective encodes either behaviour; both fall out of the
same residual.

The gain carries through to policy learning: on the same four-task suite, guidance raises H1-2 success rate from 62.3\,$\pm$\,0.4\% to 75.7\,$\pm$\,0.5\% (5 seeds, 10 episodes per task), a 13.4-point gain comparable to the G1's 11.6 points (68.1 to 79.7,
Table~ \ref{tab:necessity}) despite the different arm geometry.

The module is also generator-agnostic.
Text2HOI~\cite{cha2024text2hoi} represents the wrist pose identically to MAD-HOI, so the IK objective and the base prior transfer without modification.
Applied to its DDPM sampler as a gradient step on prediction $x_0$, the guidance lowers the IK energy on 48 of 50 ARCTIC prompts (median $2.36\to2.34$, mean $-1.3\%$) with at most $0.4$\,cm of wrist adjustment.
The same objective on the H1-2 lowers Text2HOI's energy on all 50 ARCTIC prompts (median $1.77\to1.75$, mean $-1.2\%$), again with the wrists within 0.5\,cm.
The effect is smaller than on MAD-HOI because Text2HOI's contact refiner post-processes the hands, but its direction is consistent across generators and robots.
We therefore do not depend on MAD-HOI: its checkpoint is public~\cite{madhoi}, we will release the guidance code, all generated references, and the trained policies.

\section{Conclusion and Limitations}
\label{sec:limitations}

\method{} helps a frozen HOI generator in producing motions a target humanoid can reach. The key component is running a differentiable IK solve inside its sampling loop and optimizing the robot's base.
Policies trained on the guided references succeed more often than on unguided ones (79\% vs.\ 68\% on 50 random tasks, 76\% vs.\ 62\% on the H1-2). Base placement being the larger of the two contributors (4.0 vs.\ 2.9 points) and also
the cheaper, since it adds no sampling cost.

More broadly, our comparison against post-hoc refinement
and best-of-$N$ selection is a caution for reference-generation pipelines in general: a
kinematic feasibility metric is a valid diagnostic only when the rest of the interaction is held
fixed, and driving it down directly buys feasibility at the cost of the interaction it was meant
to preserve.

Guidance is kinematic: better reachability does not imply dynamic or contact feasibility, and the RL stage absorbs what guidance does not model.
The cost is a $32\times$ slower sampler, paid once per reference; restricting guidance to a subset of the refinement steps would reduce it.

Our scope is limited in two ways.
The lower body is fixed: we optimize where the robot stands, and not how it gets there. Integrating a locomotion controller~\cite{zhang2025falcon,weng2025hdmi} is left for future work.
All evaluation is performed in simulation on ARCTIC's articulated objects; real-world rollouts are the immediate next step.
Beyond scope, fine-tuning the generator on its own guided samples would distill the guidance into the prior and remove the inference cost, at the price of one generator per embodiment.

\section*{Acknowledgments}
LLMs were used for copy-editing and debugging scripts; all technical content, experiments, and analyses are the authors' own.

\bibliographystyle{ieeetr}
\bibliography{references}

\clearpage
\appendices

\section{Implementation Details}

\subsection{Flow Matching and Energy Guidance}
\label{app:guidance}

This section expands the sampling formulation summarized in Sec.~\ref{sec:ik}, so that the
guidance derivation is self-contained for readers unfamiliar with flow matching.

\textbf{Flow-matching sampler.}
MAD-HOI~\cite{madhoi} generates continuous motion latents with a flow-matching objective.
Rather than learning a sequence of discrete denoising transitions, it learns a time-dependent
velocity field $\mathbf v_\theta(\mathbf x,t,\mathbf z)$ that transports samples from a simple
noise distribution toward the data distribution, conditioned on context $\mathbf z$. Under a
linear interpolation (coupling) path between clean latents and Gaussian noise, the learned
field defines the probability-flow ODE
\begin{equation}
\dot{\mathbf x}=\mathbf v_\theta(\mathbf x,t,\mathbf z),
\label{eq:app_pfode}
\end{equation}
whose integration from $t{=}0$ to $t{=}1$ transforms noise into a motion sample. The context
$\mathbf z$ is produced by the transformer from the currently unmasked tokens and the text
prompt.

\textbf{Energy guidance.}
Following classifier-style guidance for diffusion and flow models~\cite{dhariwal2021diffusion},
we bias sampling with an auxiliary energy $E(\mathbf x)$,
\begin{equation}
\dot{\mathbf x}=\mathbf v_\theta(\mathbf x,t,\mathbf z)-w\,\nabla_{\mathbf x}E(\mathbf x),
\label{eq:app_guidance}
\end{equation}
where $w\ge0$ controls guidance strength. The gradient term steers trajectories toward
lower-energy regions while preserving the generative prior encoded by the learned flow. In our
method $E(\mathbf x)$ is the inverse-kinematics residual of Eq.~\eqref{eq:energy}, evaluated on
the target humanoid, and guidance is applied only at inference: the generator's weights are
never updated.

We use a constant guidance weight $w{=}0.5$ rather than a schedule $w(t)$. We integrate
Eq.~\eqref{eq:app_guidance} with a fourth-order Runge--Kutta solver using 20 steps per
unmasking round. Unguided baselines are drawn with MAD-HOI's released sampler, so the
comparison in Table~\ref{tab:quality} evaluates the guided sampler as a whole (RK4 integrator
plus energy term) against the released one rather than isolating the energy term.

\textbf{Masking.}
The gradient is applied only to hand tokens, through the elementwise mask
$\mathbf m_{\mathrm{hand}}$ of Alg.~\ref{alg:outer}. Object tokens receive no guidance
gradient, so the generated object trajectory and articulation are bit-identical between a
guided and its seed-matched unguided sample. This is what makes the object-side reference,
and therefore the task reward, directly comparable across the two conditions.

\subsection{IK Optimization Details}
\label{app:ik}

\textbf{Objective.}
The guidance energy is computed from a differentiable IK optimization on the target humanoid.
The optimization combines a primary wrist-pose tracking objective with regularization terms
that improve kinematic feasibility and base consistency. Table~\ref{tab:ik_guidance} lists the
components and their weights. All weights are fixed once and shared across both embodiments,
both generators, and every task; Sec.~\ref{sec:quality} reports a sensitivity sweep over four
of them.

\begin{table}[t]
\centering
\caption{Components of the IK guidance objective. Wrist tracking is the primary task; the
remaining terms regularize toward kinematically feasible arm and base configurations. The
null-space term is a secondary optimization applied after the primary solve.}
\label{tab:ik_guidance}
\footnotesize
\setlength{\tabcolsep}{4pt}
\begin{tabular}{llp{4.6cm}}
\toprule
Component & Weight & Description \\
\midrule
Wrist position      & $w_{\mathrm{pos}}{=}250$            & Tracks the target wrist position. \\
Wrist orientation   & $w_{\mathrm{ori}}{=}5$              & Tracks the target wrist orientation. \\
Rest-pose prior     & $\lambda_{\mathrm{rest}}{=}10$      & Encourages solutions near a nominal arm configuration. \\
Joint-limit penalty & $w_{\mathrm{joint}}{=}100$          & Penalizes configurations outside joint limits. \\
Smoothness          & $w_{\mathrm{smooth}}{=}1$           & Penalizes deviation from the previous timestep's solution. \\
Root prior          & $w_{\mathrm{root}}{=}100$           & Keeps the base near the current preferred placement $\mathbf b_0$. \\
Object clearance    & $w_{\mathrm{clear}}{=}100$          & Penalizes arm configurations violating a minimum clearance radius around the object. \\
\midrule
Null-space posture  & $\lambda_{\mathrm{manip}}{=}0.03$   & Secondary objective improving manipulability without moving the wrist. \\
\bottomrule
\end{tabular}
\end{table}

\textbf{Null-space optimization.}
Wrist-pose tracking defines the primary task; the remaining terms regularize toward stable,
feasible arm and base configurations, and the clearance term discourages arm configurations
that pass close to the manipulated object. After the primary Levenberg--Marquardt solve we
apply the null-space posture objective, which adjusts the arm configuration without altering
the achieved wrist pose. This yields postures better conditioned for downstream retargeting and
policy learning, and is the component ablated as ``arm conditioning'' in
Table~\ref{tab:ablation}.

\textbf{Solver settings.}
Table~\ref{tab:ik_hyperparams} lists the optimizer parameters, which are identical across all
embodiments and tasks. A small iteration budget is sufficient for stable wrist solutions and
keeps guidance runtimes practical: the solve runs once per RK4 step, so its cost is multiplied
by 20 per unmasking round (Appendix~\ref{app:compute}).

\begin{table}[t]
\centering
\caption{Optimization parameters of the IK guidance module.}
\label{tab:ik_hyperparams}
\footnotesize
\begin{tabular}{lc}
\toprule
Parameter & Value \\
\midrule
LM iterations per timestep          & 10 \\
LM damping coefficient              & $10^{-6}$ \\
Gradient-norm termination threshold & $10^{-2}$ \\
Implicit differentiation            & enabled \\
Solve variables $\mathbf u^t$       & arm joints $\mathbf q^t$ + base translation $\mathbf b^t$ \\
Base orientation                    & fixed \\
Base-prior refresh start $s_{\mathrm{start}}$ & 0 (first unmasking step) \\
Base-prior updates per generation   & 18 (one per unmasking step) \\
\bottomrule
\end{tabular}
\end{table}

\textbf{Gradients through the solver.}
We differentiate implicitly through the LM solve in JAX, so $\partial E/\partial(\boldsymbol\tau,
\boldsymbol\phi)$ is defined without unrolling the iterations, and a JAX--PyTorch bridge
propagates the gradient through the decoder $\mathcal D$ to the latent $\mathbf x$. Implicit
differentiation keeps memory constant in the iteration count, which is what makes running the
solve inside every flow step tractable.

\subsection{Retargeting Details}
\label{app:retargeting}

Generated sequences are expressed in MANO hand parameters and object trajectories; policy
learning requires robot joint trajectories. Wrist trajectories are converted to upper-body
joint configurations with PyRoki~\cite{kim2025pyroki}, which optimizes joint states to match
target wrist positions and orientations subject to the embodiment's kinematic limits, solved
from the base placement produced by guidance rather than from a fixed origin. MANO finger
articulations are mapped to the XHand embodiment with
dex-retargeting~\cite{qin2023anyteleop}. We additionally apply contact retargeting following
DexMachina~\cite{mandi2025dexmachina} to obtain reference contact locations
$\hat{\mathbf c}_t$ on the hand links, used as auxiliary supervision during policy learning.
The generated object trajectory and articulation are used unchanged as the object-side
reference in observations, rewards, and terminations.

Because guidance never edits object tokens (Appendix~\ref{app:guidance}), the only difference
between a guided and a seed-matched unguided reference after retargeting is the arm and finger
trajectory, and the base placement they are solved from.

\subsection{RL Environment and Policy Setup}
\label{app:rl}

\textbf{Setup.}
Given retargeted trajectories, we train a policy $\pi$ to imitate the reference motion while
completing the manipulation task. We use PPO~\cite{schulman2017proximal} implemented in
rl-games~\cite{makoviichuk2021rlgames}, training in Genesis~\cite{genesis2024} on a single
NVIDIA RTX~4090 (24\,GB) with 8192 parallel environments for 2000 PPO iterations
($\approx$4\,h per policy).

\textbf{Observations.}
The policy observes short histories of robot joint positions and velocities
$(\mathbf q^{\mathrm{robot}}_t,\dot{\mathbf q}^{\mathrm{robot}}_t)$ and previous actions
$\mathbf a_t$ over the past nine steps, the normalized episode phase $\phi_t$, and sparse
future reference offsets $\hat{\mathbf q}^{\mathrm{robot}}_{t+\Delta}-\mathbf
q^{\mathrm{robot}}_t$ for $\Delta\in\{1,2,4,8,16\}$, which encourage anticipatory tracking. For
the articulated object it observes both part poses, the articulation state, and future tracking
errors against the generated object trajectory. Table~\ref{tab:observations} gives the full
specification.

\textbf{Actions.}
The policy outputs 39-D actions: 15-D \emph{delta} targets for torso and arms and 24-D
\emph{absolute} finger targets. Delta control on the upper body yields smoother large-scale
motion, while absolute finger commands improve learning given the limited joint ranges of the
XHand. Action parameterization details --- delta scale, PLAI interpolation coefficient, and
action delay --- are listed in Table~\ref{tab:hyperparams}.

\textbf{Rewards.}
The total reward is
\begin{equation}
\begin{aligned}
r_t ={}& w_{\mathrm{task}}r_{\mathrm{task}}
        +w_{\mathrm{bc}}r_{\mathrm{imi,upper}}\\
      &+w_{\mathrm{finger}}r_{\mathrm{imi,dex}}
        +w_{\mathrm{contact}}r_{\mathrm{contact}}\\
      &-\lambda_{\mathrm{act}}r_{\mathrm{act}}
        -\lambda_{\mathrm{rate}}r_{\mathrm{act\text{-}rate}}\\
      &-\lambda_{\mathrm{vel}}r_{\mathrm{vel}}
        -\lambda_{\mathrm{acc}}r_{\mathrm{acc}}
        -\lambda_{\mathrm{ctrl}}r_{\mathrm{ctrl}},
\end{aligned}
\label{eq:app_reward}
\end{equation}
where $r_{\mathrm{task}}$ rewards agreement with the reference object pose and articulation,
$r_{\mathrm{imi,upper}}$ and $r_{\mathrm{imi,dex}}$ are the decoupled arm and hand imitation
streams of Sec.~\ref{sec:rl}, and $r_{\mathrm{contact}}$ rewards agreement with the retargeted
contact locations. The remaining terms regularize actions, control effort, and smoothness.
Table~\ref{tab:reward_terms} defines each term and Table~\ref{tab:hyperparams} gives all
weights and temperatures.

\textbf{Terminations.}
Episodes end at the task-dependent horizon $T_{\max}$, on NaN in the simulation state, when the
object falls below the table, when object articulation departs the reference by more than
$\epsilon_a$, or when either imitation stream exceeds its threshold --- the \emph{maximum} over
arm and torso joints for the upper body, the \emph{mean} over finger joints for the hand.
Table~\ref{tab:termination_conditions} states each condition; thresholds are applied to the
half-squared joint error $\tfrac12\Delta q^2$, so $\epsilon_{\mathrm{upper}}{=}0.12$
corresponds to a $0.49$\,rad deviation on the worst joint.

\textbf{Curricula.}
We never actuate the object, so unlike DexMachina~\cite{mandi2025dexmachina} we use no
virtual-object-controller curriculum and no reward curriculum.
We have an additional entropy curriculum, which halves the entropy
\% coefficient and disables random state initialization once mean episode length exceeds 80\% of the horizon). Without this curriculum, the policy can learn specific segments of the demonstration (lifting box, opening box), but fails to join the different segments into a single demonstration. 

\input{appendix_tables}

\section{Additional Quantitative Results}

\subsection{Kinematic Analysis}
\label{app:kinematics}

\begin{figure}[t]
    \centering
    \includegraphics[width=\columnwidth]{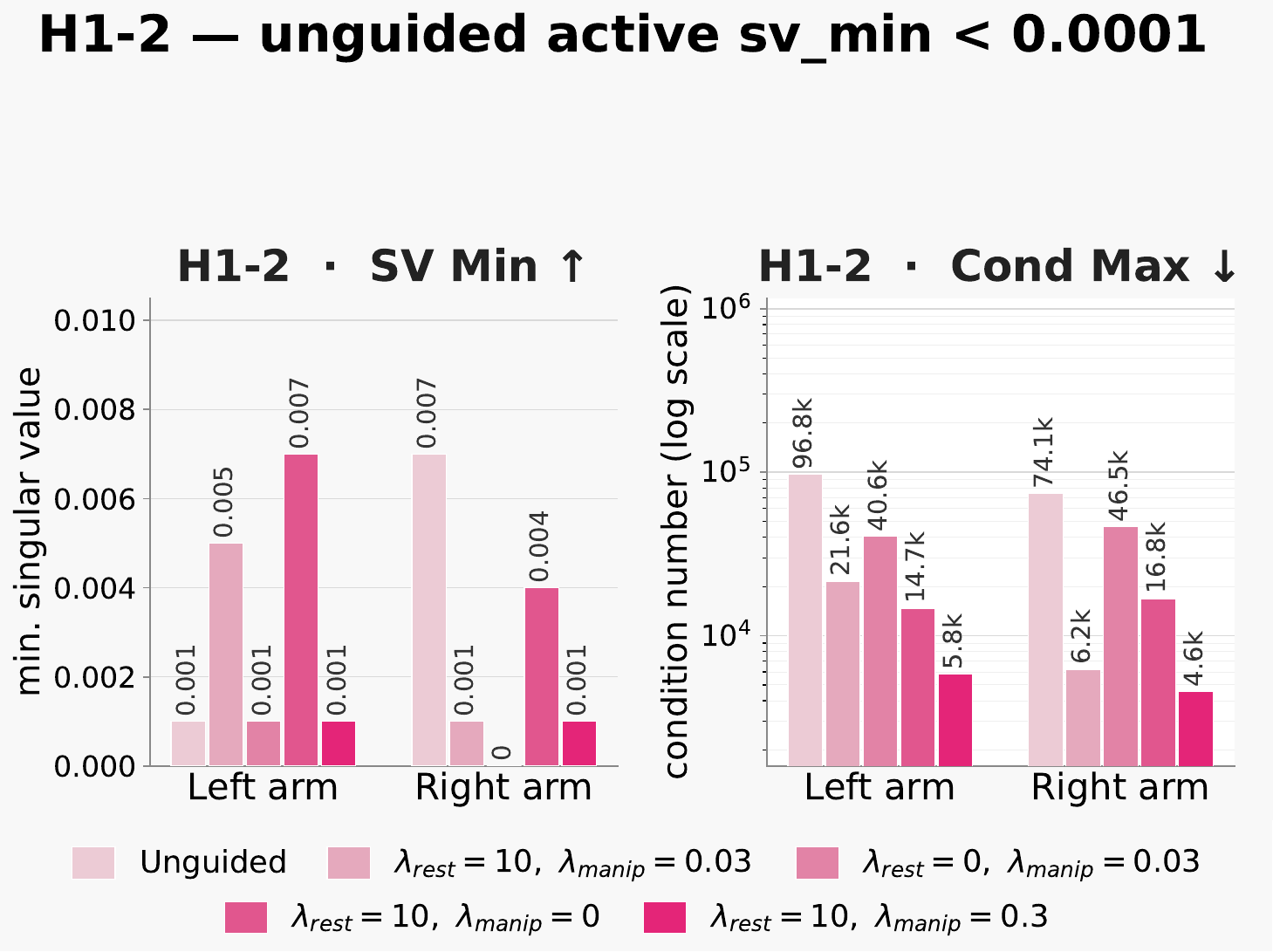}
    \caption{Worst-case conditioning of the Unitree H1-2 per arm over the ARCTIC prompts:
    minimum Jacobian singular value ($\uparrow$) and maximum condition number ($\downarrow$,
    log scale). Compare Fig.~\ref{fig:singgrid} for the G1.}
    \label{fig:h1_singularity}
\end{figure}

\textbf{Conditioning on the H1-2.}
Figure~\ref{fig:h1_singularity} reports the minimum Jacobian singular value and maximum
condition number for the Unitree H1-2 under the protocol of Sec.~\ref{sec:quality}. The trends
are less pronounced than on the G1: because the H1-2's 68.5\,cm arms are closer to human
proportions, both guided and unguided conditions remain far from singularity
($\svmin\approx0.13$) and there are few near-singular configurations for base placement to
remove. The benefit on this embodiment therefore comes from the residual reachability error
rather than from singularity avoidance, which is consistent with the downstream gain reported
in Sec.~\ref{sec:cross}.

\textbf{Per-setting weight sweep.}
Sec.~\ref{sec:quality} reports that perturbing the IK weights moves the median IK energy by at
most 0.03 and $\svmin$ by less than 1\%, while every perturbation degrades generation quality
(FID 0.615--0.869 vs.\ 0.545). Table~\ref{tab:weight_sweep} gives the per-setting numbers behind those ranges.

\begin{table}[t]
\centering
\caption{Sensitivity to IK objective weights on the same 400 ARCTIC test prompts. The default
setting is the one used in all other experiments and was fixed before this sweep was run.}
\label{tab:weight_sweep}
\footnotesize
\setlength{\tabcolsep}{4pt}
\begin{tabular}{lcccc}
\toprule
Setting & Median $E\downarrow$ & $\svmin\uparrow$ & FID$\downarrow$ & Top-3$\uparrow$ \\
\midrule
Default (ours)                           & 1.81 & 0.082 & \textbf{0.545} & \textbf{0.872} \\
$\lambda_{\mathrm{rest}}{=}0$            & 2.05 & 0.054 & 0.615 & 0.869 \\
$\lambda_{\mathrm{manip}}{=}0$           & 1.94 & 0.082 & 0.650 & 0.861 \\
$\lambda_{\mathrm{manip}}{\times}3$      & 1.78 & 0.086 & 0.869 & 0.835 \\
$w_{\mathrm{root}}{\times}0.3$           & 1.88 & 0.083 & 0.810 & 0.830 \\
\bottomrule
\end{tabular}
\end{table}

\subsection{Additional Downstream Results}
\label{app:downstream}

\textbf{Stratified comparison of guidance and selection.}
Sec.~\ref{sec:necessity} states that best-of-$N$ selection reaches lower IK energy than
guidance on 88--94\% of prompts regardless of difficulty. Table~\ref{tab:stratified} gives the
per-bin numbers. Prompts are split into terciles by the IK energy of the \emph{unguided,
optimized-base} sample, so the binning does not depend on the method under test. The lower
block reports the two quantities that explain the pattern: sample-to-sample variance of $E$
within a prompt, against the size of the shift guidance applies. Selection exploits the former
and guidance the latter, and the gap between them widens with difficulty --- which is why
selection's energy advantage does not shrink on hard prompts even though its downstream
advantage never appears at all.

\begin{table}[t]
\centering
\caption{Selection's energy advantage does not depend on difficulty. 51 prompts split into
\% terciles by unguided optimized-base $E$ (bin edges $E<$1.80, 1.80--3.00, $>$3.00;
\% $n{=}17$ each).}
\label{tab:stratified}
\setlength{\tabcolsep}{2pt}
\renewcommand{\arraystretch}{1.1}
\begin{tabularx}{\columnwidth}{@{}l*{5}{>{\centering\arraybackslash}X}@{}}
\toprule
& \multicolumn{3}{c}{\shortstack{\% Reaching lower $E$\\than EmbodyHOI}}
& \multicolumn{2}{c}{} \\
\cmidrule(lr){2-6}
Method & Easy & Med. & Hard & Median $E\downarrow$ & SR $\%\uparrow$ \\
\midrule
Best-of-8 by $E$  & 65\% & 88\% & 75\% & 1.36 & 73.1\% \\
Best-of-32 by $E$ & 88\% & 94\% & 88\% & 1.21 & 74.0\% \\
EmbodyHOI (ref.)  & --   & --   & --   & 1.81 & 79.7\% \\
\bottomrule
\end{tabularx}
\end{table}


\textbf{Where the downstream gain concentrates.}
Table~\ref{tab:sr_stratified} splits the 50 randomly sampled tasks by the conditioning of
their unguided reference. The subset boundary ($\svmin{<}0.1$) is fixed before any policy is
trained. Most randomly drawn prompts are already reachable and guidance changes little on
them; the aggregate gain is carried by the poorly conditioned minority, which is the
population the method is designed for.

\begin{table}[t]
\centering
\caption{Success rate on the 50 randomly sampled G1 tasks, split by $\svmin$ of the unguided
reference; mean $\pm$ s.e.m.\ over 5 seeds.}
\label{tab:sr_stratified}
\footnotesize
\begin{tabular}{lccc}
\toprule
Subset & $n$ & Unguided SR\,\% & Guided SR\,\%$\uparrow$ \\
\midrule
All tasks                         & 50 & 68.1 & \textbf{79.2} \\
Well-conditioned ($\svmin\ge0.1$) & 40 & 70.7 & \textbf{78.3} \\
Hard ($\svmin<0.1$)               & 10 & 57.9 & \textbf{82.1} \\
\bottomrule
\end{tabular}
\end{table}

\textbf{Reachability-free reference point.}
Sec.~\ref{sec:downstream} notes that DexMachina's released floating-wrist system reaches 86.1\%
on the four-task suite. That system commands a free-floating 6-DoF wrist with no arm chain, so
it never encounters the reachability problem we study; its number is an upper bound obtained by
assuming away the constraint, not a baseline. We report it here so the gap to our 79.3\% can be
read as the cost of driving a real 15-DoF arm rather than as a deficiency of the RL recipe.

\section{Computational Cost}
\label{app:compute}

Table~\ref{tab:runtime} summarizes the cost of each pipeline stage. Guidance is applied only at
inference and adds no training cost to the generative model. The $32\times$ increase in sampling
time is the method's main practical cost; it is paid once per reference and amortized over the
$\approx$4\,h of policy training that reference supports, and it remains far cheaper than
training an embodiment-specific motion generator, which additionally requires robot-feasible
data for every new robot (Sec.~\ref{sec:related}).

\begin{table}[t]
\centering
\caption{Approximate runtime per pipeline stage. Generation was measured on an NVIDIA A100 and
PPO training on an RTX~4090. Runtimes are approximate and vary with sequence length and task.}
\label{tab:runtime}
\footnotesize
\begin{tabular}{lcc}
\toprule
Stage & Runtime & Hardware \\
\midrule
MAD-HOI generation (unguided)  & 5\,s          & A100 \\
IK-guided generation (ours)    & 160\,s        & A100 \\
Post-hoc IK refinement         & 55\,s        & A100 \\
Best-of-32 selection           & 160\,s        & A100 \\
Retargeting per sequence       & 50\,s        & CPU \\
PPO training per policy        & $\approx$4\,h & RTX~4090 \\
\bottomrule
\end{tabular}
\end{table}

Restricting guidance to a subset of unmasking rounds, or to a subset of RK4 steps within each
round, would reduce the sampling cost proportionally; we did not explore this tradeoff and note
it as the most direct route to cheaper guidance (Sec.~\ref{sec:limitations}).

\section{Additional Qualitative Results}
\label{app:qualitative}

\begin{figure}[!tb]
    \centering
    \includegraphics[width=\columnwidth]{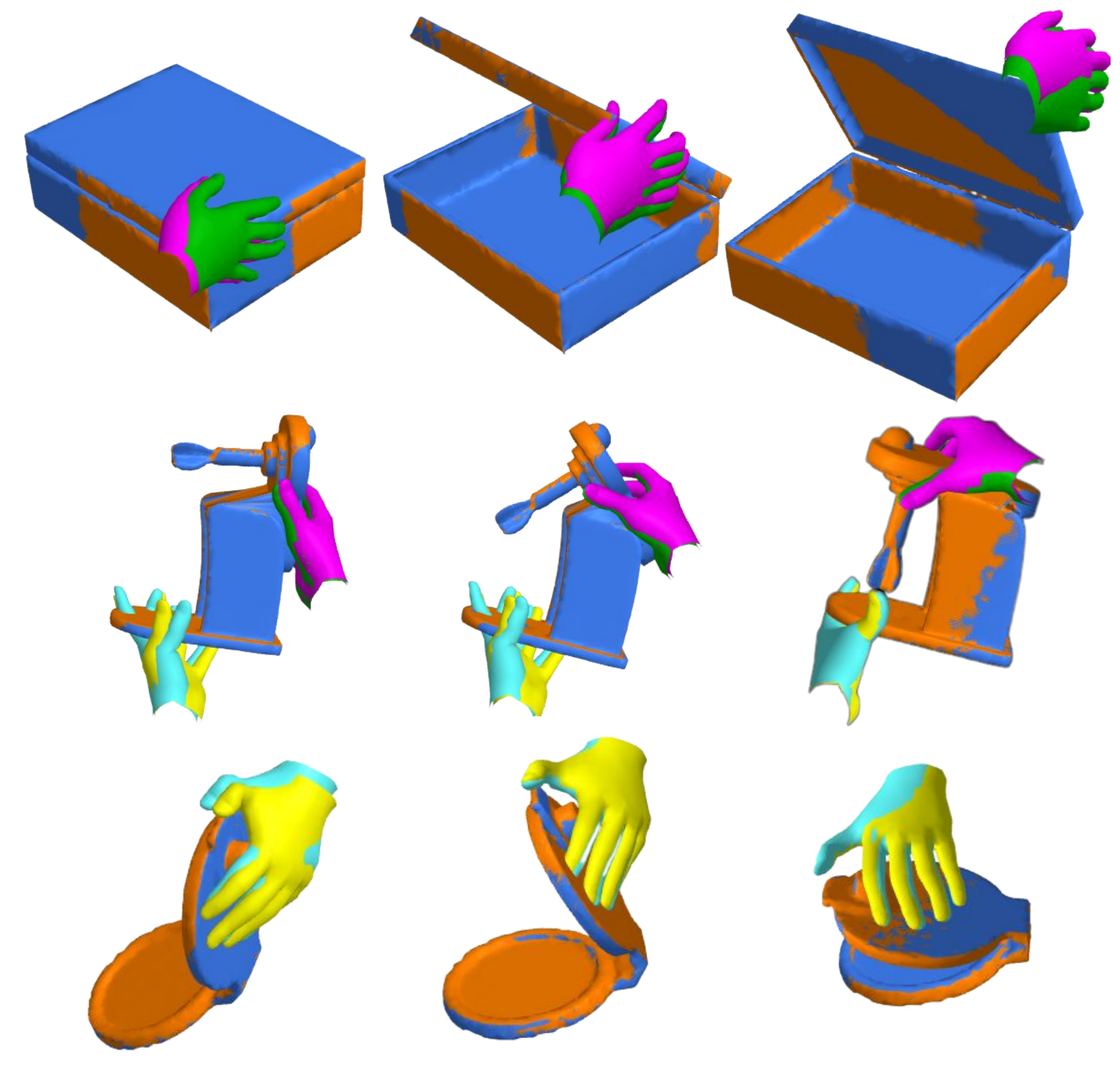}
    \caption{Unguided MAD-HOI generations (left: cyan; right: green) against our IK-guided
    generations (left: yellow; right: magenta) for matched prompts and seeds. Guidance steers
    wrist trajectories toward kinematically feasible configurations for the target embodiment
    while preserving the structure of the hand--object interaction.}
    \label{fig:app_guided_vs_unguided}
\end{figure}

\textbf{Guided versus unguided generations.}
Figure~\ref{fig:app_guided_vs_unguided} compares unguided MAD-HOI samples with our IK-guided
samples for the same prompt and seed. Guidance reshapes wrist trajectories toward configurations
the target humanoid can reach while preserving the semantic structure of the interaction ---
the grasp type, the contact locations, and the object motion are unchanged, consistent with the
unchanged contact ratio and penetration in Table~\ref{tab:quality}. This is the static
counterpart to the claim in Sec.~\ref{sec:quality} that guidance does not collapse the semantic
content of the interaction.

\begin{figure*}[t]
    \centering
    \includegraphics[width=\textwidth]{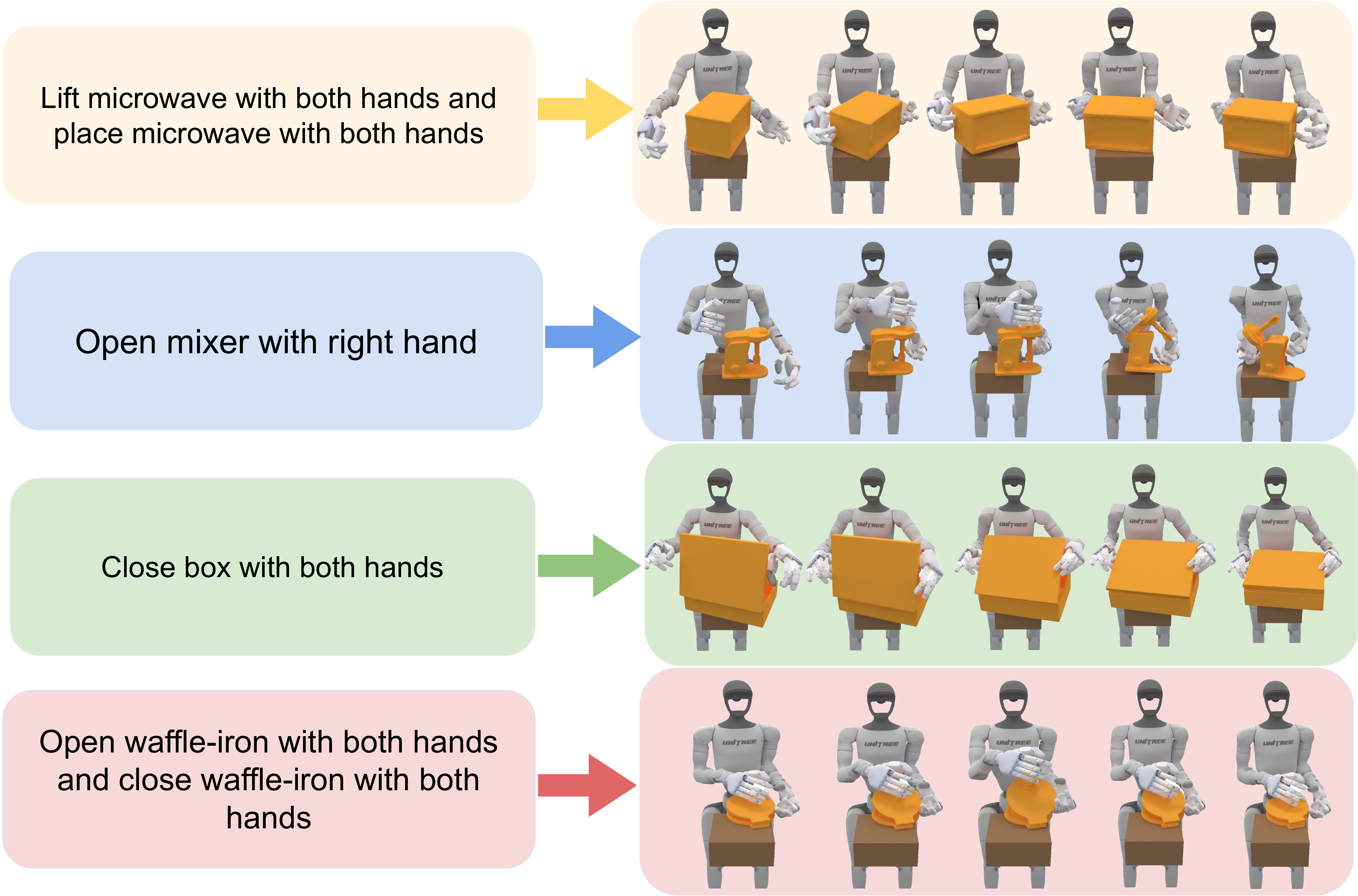}
    \caption{Additional articulated-object manipulation tasks executed on the Unitree G1 by
    policies trained from \method{} references, spanning opening, closing, lifting, and
    bimanual behaviors across the ARCTIC object set.}
    \label{fig:app_more_samples}
\end{figure*}

\textbf{Additional executed tasks.}
See Figure \ref{fig:app_more_samples} for more samples generated from our pipeline.

\end{document}

%% file: appendix_tables.tex
%

\begin{table*}[t]
\centering
\small
\setlength{\tabcolsep}{10pt}
\renewcommand{\arraystretch}{1.2}
\begin{tabular}{p{0.40\textwidth} p{0.50\textwidth}}
\toprule
\textbf{Observation} & \textbf{Interpretation} \\
\midrule
$\{\mathbf q^{\mathrm{robot}}_{t-k}\}_{k=0}^{8}$ &
Robot joint positions over the past 9 steps (proprioception). \\

$\{\dot{\mathbf q}^{\mathrm{robot}}_{t-k}\}_{k=0}^{8}$ &
Robot joint velocities over the past 9 steps. \\

$\{\mathbf a_{t-k}\}_{k=0}^{8}$ &
Previous actions over the past 9 steps ($n_a{=}39$ controlled DoFs). \\

$\phi_t$ &
Normalized episode phase $2t/T_{\max}-1$. \\

$\hat{\mathbf q}^{\mathrm{robot}}_{t+\Delta}-\mathbf q^{\mathrm{robot}}_t,\quad
\Delta\in\{1,2,4,8,16\}$ &
Future reference joint offsets (retargeted trajectory minus current joints). \\
\midrule

$T^{\mathrm{obj}}_{t,k},\ R^{\mathrm{obj}}_{t,k},\quad k\in\{1,2\}$ &
3D position and 4D orientation of both articulated object parts. \\

$\alpha^{\mathrm{obj}}_t$ &
1D object articulation (joint opening). \\

$d_p,\ d_r,\ (\alpha^{\mathrm{obj}}_{t+h}-\alpha^{\mathrm{obj}}_t),\quad h\in\{1,2,4,8,16\}$ &
Future tracking errors against the reference object pose
($d_p{=}\|T^{\mathrm{obj}}_t-\hat T^{\mathrm{obj}}_{t+h}\|_2$,
$d_r{=}d_{\mathrm{rot}}(R^{\mathrm{obj}}_t,\hat R^{\mathrm{obj}}_{t+h})$). \\

$\hat{\mathbf p}^{\mathrm{obj}}_{t+h},\quad h\in\{1,2,4,8,16\}$ &
Reference object root positions at future timesteps. \\

$\hat{\mathbf p}^{\mathrm{obj}}_t-\mathbf p^{\mathrm{obj}}_t$ &
Current object root position error against the reference. \\
\bottomrule
\end{tabular}
\caption{Policy observations for bimanual manipulation on articulated objects. The reference
quantities $\hat\cdot$ come from the generated (guided or unguided) trajectory, never from
ground truth.}
\label{tab:observations}
\end{table*}

\begin{table*}[t]
\centering
\small
\setlength{\tabcolsep}{10pt}
\renewcommand{\arraystretch}{1.2}
\begin{tabular}{p{0.40\textwidth} p{0.50\textwidth}}
\toprule
\textbf{Reward term} & \textbf{Interpretation} \\
\midrule
$r_{\mathrm{task}}=
\begin{aligned}[t]
&\left\|T^{\mathrm{obj}}_t-\hat T^{\mathrm{obj}}_t\right\|_2\cdot
 d_{\mathrm{rot}}(R^{\mathrm{obj}}_t,\hat R^{\mathrm{obj}}_t)\\
&\cdot-\beta_a\left(\alpha^{\mathrm{obj}}_t-\hat\alpha^{\mathrm{obj}}_t\right)^2
\end{aligned}$ &
Rewards matching the reference object position, orientation, and articulation in the world
frame. \\

$r_{\mathrm{imi,upper}}=\left\|\mathbf q^{\mathrm{upper}}_t-\hat{\mathbf q}^{\mathrm{upper}}_t\right\|_2^2$ &
Upper-body imitation stream: tracking of the 15 torso and arm joints. \\

$r_{\mathrm{imi,dex}}=\left\|\mathbf q^{\mathrm{dex}}_t-\hat{\mathbf q}^{\mathrm{dex}}_t\right\|_2^2$ &
Dexterous imitation stream: tracking of the 24 finger joints. \\

$r_{\mathrm{contact}}=\left\|\mathbf c_t-\hat{\mathbf c}_t\right\|_2$ &
Rewards matching simulated contact locations to the retargeted reference contacts on hand
links. \\
\midrule

$r_{\mathrm{act}}=\left\|\mathbf a_t\right\|_2$ & Penalizes large actions. \\
$r_{\mathrm{act\text{-}rate}}=\left\|\mathbf a_t-\mathbf a_{t-1}\right\|_2^2$ & Penalizes large action rates. \\
$r_{\mathrm{ctrl}}=\left\|\boldsymbol\tau_t\right\|_1$ & Penalizes large joint control torques. \\
$r_{\mathrm{vel}}=\left\|\dot{\mathbf q}_t\right\|_1$ & Penalizes large joint velocities. \\
$r_{\mathrm{acc}}=\left\|\ddot{\mathbf q}_t\right\|_1$ & Penalizes large joint accelerations. \\
\bottomrule
\end{tabular}
\caption{Reward terms. The two imitation streams are kept separate rather than summed over all
39 DoFs; Table~\ref{tab:ablation} shows this decoupling is worth 5.2 points of success rate.}
\label{tab:reward_terms}
\end{table*}

\begin{table*}[t]
\centering
\small
\setlength{\tabcolsep}{10pt}
\renewcommand{\arraystretch}{1.2}
\begin{tabular}{p{0.40\textwidth} p{0.50\textwidth}}
\toprule
\textbf{Termination condition} & \textbf{Interpretation} \\
\midrule
$t\ge T_{\max}$ & Ends the episode at the task-dependent horizon. \\

$\mathrm{NaN}$ in simulation state & Ends the episode on numerical failure in robot or object state. \\

$\max_i\ \tfrac12\left(q^{\mathrm{upper}}_{t,i}-\hat q^{\mathrm{upper}}_{t,i}\right)^2>\epsilon_{\mathrm{upper}}$ &
Ends the episode when \emph{any} arm or torso joint deviates too far: a large arm failure ends
the rollout. \\

$\mathrm{mean}_i\ \tfrac12\left(q^{\mathrm{dex}}_{t,i}-\hat q^{\mathrm{dex}}_{t,i}\right)^2>\epsilon_{\mathrm{dex}}$ &
Ends the episode on \emph{average} finger deviation: an isolated finger error is not
catastrophic. \\

$p^{\mathrm{obj}}_{t,z}<h_{\mathrm{table}}$ & Ends the episode when the object falls below the table. \\

$\tfrac12\left(\alpha^{\mathrm{obj}}_t-\hat\alpha^{\mathrm{obj}}_t\right)^2>\epsilon_a$ &
Ends the episode when object articulation departs the reference trajectory. \\
\bottomrule
\end{tabular}
\caption{Episode termination conditions. Thresholds are applied to the half-squared joint
error, so $\epsilon_{\mathrm{upper}}{=}0.12$ corresponds to a $0.49$\,rad deviation on the worst
joint. The asymmetry between the maximum (arm) and the mean (hand) is the termination-side
counterpart of the decoupled rewards.}
\label{tab:termination_conditions}
\end{table*}

\begin{table*}[!ht]
\centering
\small
\setlength{\tabcolsep}{8pt}
\renewcommand{\arraystretch}{1.15}
\begin{tabular}{llc}
\toprule
\textbf{Category} & \textbf{Parameter} & \textbf{Value} \\
\midrule
\multicolumn{3}{l}{\textit{PPO / training}} \\
& Parallel environments $B$        & 8192 \\
& PPO iterations                   & 2000 \\
& Rollout horizon (steps per env)  & 16 \\
& Learning rate                    & $3{\times}10^{-4}$ \\
& Discount $\gamma$ / GAE $\tau$   & 0.99 / 0.95 \\
& PPO clip $\epsilon$              & 0.2 \\
& Entropy coefficient              & 0.0055 \\
& Bounds-loss coefficient          & $10^{-4}$ \\
& Minibatch size                   & $8B$ \\
& PPO epochs per rollout           & $\max(1,\lfloor 5B/4096\rfloor)$ \\
& Reward-shaper scale              & 0.1 \\
& Policy MLP                       & 512, 512, 256, 128 (ELU) \\
& Training hardware                & 1$\times$ RTX~4090 (24\,GB) \\
\midrule
\multicolumn{3}{l}{\textit{Environment}} \\
& Episode length $T_{\max}$        & task dependent \\
& Object demo horizons             & $\{1,2,4,8,16\}$ \\
& Robot history steps              & $\{0,\ldots,8\}$ \\
& Robot future steps               & $\{1,2,4,8,16\}$ \\
\midrule
\multicolumn{3}{l}{\textit{Reward weights}} \\
& $w_{\mathrm{task}}$              & 1.3 \\
& $w_{\mathrm{bc}}$ (upper body)   & 1.0 \\
& $w_{\mathrm{finger}}$            & 0.15 \\
& $w_{\mathrm{contact}}$           & 0.7 \\
\midrule
\multicolumn{3}{l}{\textit{Reward shaping}} \\
& $\beta_p,\ \beta_r,\ \beta_a$    & 10, 1, 5 \\
& $\beta_{\mathrm{bc}}$ (upper)    & 30 \\
& $\beta_{\mathrm{finger}}$        & 20 \\
& $\beta_{\mathrm{contact}}$       & 30 \\
\midrule
\multicolumn{3}{l}{\textit{Penalties}} \\
& $\lambda_{\mathrm{act}}$         & 0.02 \\
& $\lambda_{\mathrm{rate}}$        & 0.003 \\
& $\lambda_{\mathrm{vel}}$         & 0.03 \\
& $\lambda_{\mathrm{acc}}$         & 0.025 \\
& $\lambda_{\mathrm{ctrl}}$        & 0.01 \\
\midrule
\multicolumn{3}{l}{\textit{Termination thresholds} (on $\tfrac12\Delta q^2$)} \\
& $\epsilon_{\mathrm{upper}}$      & 0.12 \\
& $\epsilon_{\mathrm{dex}}$        & 0.2 \\
& $\epsilon_a$ (articulation)      & 0.4 \\
\midrule
\multicolumn{3}{l}{\textit{Action parameterization}} \\
& Upper body (15 DoF)              & delta (PLAI) \\
& Fingers (24 DoF)                 & absolute \\
& Joint delta scale                & 0.05 \\
& PLAI $\alpha$                    & $[0.8, 1.0]$ \\
& Action delay (steps)             & $[0, 2]$ \\
\bottomrule
\end{tabular}
\caption{RL hyperparameters. The same values are used for every condition in
Tables~\ref{tab:necessity} and~\ref{tab:ablation} and for both embodiments; only the reference
trajectory differs between conditions. Domain randomization and observation/action noise are
omitted.}
\label{tab:hyperparams}
\end{table*}

%% file: references.bib
@inproceedings{shaw2024leap,
  title={LEAP hand v2: Dexterous, low-cost anthropomorphic hybrid rigid soft hand for robot learning},
  author={Shaw, Kenneth and Pathak, Deepak},
  booktitle={2nd Workshop on Dexterous Manipulation: Design, Perception and Control (RSS)},
  year={2024}
}

@article{qiu2025humanoid,
  title={Humanoid policy\~{} human policy},
  author={Qiu, Ri-Zhao and Yang, Shiqi and Cheng, Xuxin and Chawla, Chaitanya and Li, Jialong and He, Tairan and Yan, Ge and Yoon, David J and Hoque, Ryan and Paulsen, Lars and others},
  journal={arXiv preprint arXiv:2503.13441},
  year={2025}
}

@inproceedings{fan2023arctic,
  title={ARCTIC: A dataset for dexterous bimanual hand-object manipulation},
  author={Fan, Zicong and Taheri, Omid and Tzionas, Dimitrios and Kocabas, Muhammed and Kaufmann, Manuel and Black, Michael J and Hilliges, Otmar},
  booktitle={Proceedings of the IEEE/CVF conference on computer vision and pattern recognition},
  pages={12943--12954},
  year={2023}
}

@inproceedings{taheri2020grab,
  title={GRAB: A dataset of whole-body human grasping of objects},
  author={Taheri, Omid and Ghorbani, Nima and Black, Michael J and Tzionas, Dimitrios},
  booktitle={European conference on computer vision},
  pages={581--600},
  year={2020},
  organization={Springer}
}

@article{mandi2025dexmachina,
  title={Dexmachina: Functional retargeting for bimanual dexterous manipulation},
  author={Mandi, Zhao and Hou, Yifan and Fox, Dieter and Narang, Yashraj and Mandlekar, Ajay and Song, Shuran},
  journal={arXiv preprint arXiv:2505.24853},
  year={2025}
}

@article{kalaria2025dreamcontrol,
  title={Dreamcontrol: Human-inspired whole-body humanoid control for scene interaction via guided diffusion},
  author={Kalaria, Dvij and Harithas, Sudarshan S and Katara, Pushkal and Kwak, Sangkyung and Bhagat, Sarthak and Sastry, Shankar and Sridhar, Srinath and Vemprala, Sai and Kapoor, Ashish and Huang, Jonathan Chung-Kuan},
  journal={arXiv preprint arXiv:2509.14353},
  year={2025}
}

@article{hsieh2025dexman,
  title={DexMan: Learning Bimanual Dexterous Manipulation from Human and Generated Videos},
  author={Hsieh, Jhen and Tu, Kuan-Hsun and Hung, Kuo-Han and Ke, Tsung-Wei},
  journal={arXiv preprint arXiv:2510.08475},
  year={2025}
}

@article{harithas2026dreamcontrol,
  title={DreamControl-v2: Simpler and Scalable Autonomous Humanoid Skills via Trainable Guided Diffusion Priors},
  author={Harithas, Sudarshan and Kwak, Sangkyung and Katara, Pushkal and Deolasee, Srujan and Kalaria, Dvij and Sridhar, Srinath and Vemprala, Sai and Kapoor, Ashish and Huang, Jonathan Chung-Kuan},
  journal={arXiv preprint arXiv:2604.00202},
  year={2026}
}

@article{peng2018deepmimic,
  title={Deepmimic: Example-guided deep reinforcement learning of physics-based character skills},
  author={Peng, Xue Bin and Abbeel, Pieter and Levine, Sergey and Van de Panne, Michiel},
  journal={ACM Transactions On Graphics (TOG)},
  volume={37},
  number={4},
  pages={1--14},
  year={2018},
  publisher={ACM New York, NY, USA}
}

@article{peng2021amp,
  title={Amp: Adversarial motion priors for stylized physics-based character control},
  author={Peng, Xue Bin and Ma, Ze and Abbeel, Pieter and Levine, Sergey and Kanazawa, Angjoo},
  journal={ACM Transactions on Graphics (ToG)},
  volume={40},
  number={4},
  pages={1--20},
  year={2021},
  publisher={ACM New York, NY, USA}
}

@article{wang2023physhoi,
  title={Physhoi: Physics-based imitation of dynamic human-object interaction},
  author={Wang, Yinhuai and Lin, Jing and Zeng, Ailing and Luo, Zhengyi and Zhang, Jian and Zhang, Lei},
  journal={arXiv preprint arXiv:2312.04393},
  year={2023}
}

@article{luo2025sonic,
  title={Sonic: Supersizing motion tracking for natural humanoid whole-body control},
  author={Luo, Zhengyi and Yuan, Ye and Wang, Tingwu and Li, Chenran and Chen, Sirui and Castaneda, Fernando and Cao, Zi-Ang and Li, Jiefeng and Minor, David and Ben, Qingwei and others},
  journal={arXiv preprint arXiv:2511.07820},
  year={2025}
}

@article{he2024omnih2o,
  title={Omnih2o: Universal and dexterous human-to-humanoid whole-body teleoperation and learning},
  author={He, Tairan and Luo, Zhengyi and He, Xialin and Xiao, Wenli and Zhang, Chong and Zhang, Weinan and Kitani, Kris and Liu, Changliu and Shi, Guanya},
  journal={arXiv preprint arXiv:2406.08858},
  year={2024}
}

@inproceedings{kim2025pyroki,
  title={Pyroki: A modular toolkit for robot kinematic optimization},
  author={Kim, Chung Min and Yi, Brent and Choi, Hongsuk and Ma, Yi and Goldberg, Ken and Kanazawa, Angjoo},
  booktitle={2025 IEEE/RSJ International Conference on Intelligent Robots and Systems (IROS)},
  pages={1312--1319},
  year={2025},
  organization={IEEE}
}

@inproceedings{li2025maniptrans,
  title={Maniptrans: Efficient dexterous bimanual manipulation transfer via residual learning},
  author={Li, Kailin and Li, Puhao and Liu, Tengyu and Li, Yuyang and Huang, Siyuan},
  booktitle={Proceedings of the IEEE/CVF Conference on Computer Vision and Pattern Recognition},
  pages={6991--7003},
  year={2025}
}

@article{pan2025spider,
  title={Spider: Scalable physics-informed dexterous retargeting},
  author={Pan, Chaoyi and Wang, Changhao and Qi, Haozhi and Liu, Zixi and Bharadhwaj, Homanga and Sharma, Akash and Wu, Tingfan and Shi, Guanya and Malik, Jitendra and Hogan, Francois},
  journal={arXiv preprint arXiv:2511.09484},
  year={2025}
}

@article{zhang2025falcon,
  title={Falcon: Learning force-adaptive humanoid loco-manipulation},
  author={Zhang, Yuanhang and Yuan, Yifu and Gurunath, Prajwal and Gupta, Ishita and Omidshafiei, Shayegan and Agha-mohammadi, Ali-akbar and Vazquez-Chanlatte, Marcell and Pedersen, Liam and He, Tairan and Shi, Guanya},
  journal={arXiv preprint arXiv:2505.06776},
  year={2025}
}

@article{weng2025hdmi,
  title={Hdmi: Learning interactive humanoid whole-body control from human videos},
  author={Weng, Haoyang and Li, Yitang and Sobanbabu, Nikhil and Wang, Zihan and Luo, Zhengyi and He, Tairan and Ramanan, Deva and Shi, Guanya},
  journal={arXiv preprint arXiv:2509.16757},
  year={2025}
}

@inproceedings{he2025hover,
  title={Hover: Versatile neural whole-body controller for humanoid robots},
  author={He, Tairan and Xiao, Wenli and Lin, Toru and Luo, Zhengyi and Xu, Zhenjia and Jiang, Zhenyu and Kautz, Jan and Liu, Changliu and Shi, Guanya and Wang, Xiaolong and others},
  booktitle={2025 IEEE International Conference on Robotics and Automation (ICRA)},
  pages={9989--9996},
  year={2025},
  organization={IEEE}
}

@article{liao2025beyondmimic,
  title={Beyondmimic: From motion tracking to versatile humanoid control via guided diffusion},
  author={Liao, Qiayuan and Truong, Takara E and Huang, Xiaoyu and Gao, Yuman and Tevet, Guy and Sreenath, Koushil and Liu, C Karen},
  journal={arXiv preprint arXiv:2508.08241},
  year={2025}
}

@article{yuan2025hermes,
  title={Hermes: Human-to-robot embodied learning from multi-source motion data for mobile dexterous manipulation},
  author={Yuan, Zhecheng and Wei, Tianming and Gu, Langzhe and Hua, Pu and Liang, Tianhai and Chen, Yuanpei and Xu, Huazhe},
  journal={arXiv preprint arXiv:2508.20085},
  year={2025}
}

@article{qin2023anyteleop,
  title={Anyteleop: A general vision-based dexterous robot arm-hand teleoperation system},
  author={Qin, Yuzhe and Yang, Wei and Huang, Binghao and Van Wyk, Karl and Su, Hao and Wang, Xiaolong and Chao, Yu-Wei and Fox, Dieter},
  journal={arXiv preprint arXiv:2307.04577},
  year={2023}
}

@inproceedings{xie2024omnicontrol,
  title={Omnicontrol: Control any joint at any time for human motion generation},
  author={Xie, Yiming and Jampani, Varun and Zhong, Lei and Sun, Deqing and Jiang, Huaizu},
  booktitle={International Conference on Learning Representations},
  volume={2024},
  pages={28176--28194},
  year={2024}
}

@article{park2025demodiffusion,
  title={Demodiffusion: One-shot human imitation using pre-trained diffusion policy},
  author={Park, Sungjae and Bharadhwaj, Homanga and Tulsiani, Shubham},
  journal={arXiv preprint arXiv:2506.20668},
  year={2025}
}

@article{luo2024omnigrasp,
  title={Omnigrasp: Grasping diverse objects with simulated humanoids},
  author={Luo, Zhengyi and Cao, Jinkun and Christen, Sammy and Winkler, Alexander and Kitani, Kris and Xu, Weipeng},
  journal={Advances in Neural Information Processing Systems},
  volume={37},
  pages={2161--2184},
  year={2024}
}

@inproceedings{tevet2023mdm,
  title     = {Human Motion Diffusion Model},
  author    = {Tevet, Guy and Raab, Sigal and Gordon, Brian and Shafir, Yonatan and Cohen-Or, Daniel and Bermano, Amit H.},
  booktitle = {International Conference on Learning Representations (ICLR)},
  year      = {2023}
}

@article{zhang2022motiondiffuse,
  title   = {MotionDiffuse: Text-Driven Human Motion Generation with Diffusion Model},
  author  = {Zhang, Mingyuan and Cai, Zhongang and Pan, Liang and Hong, Fangzhou and Guo, Xinying and Yang, Lei and Liu, Ziwei},
  journal = {arXiv preprint arXiv:2208.15001},
  year    = {2022}
}

@inproceedings{guo2024momask,
  title     = {{MoMask}: Generative Masked Modeling of {3D} Human Motions},
  author    = {Guo, Chuan and Mu, Yuxuan and Javed, Muhammad Gohar and Wang, Sen and Cheng, Li},
  booktitle = {Proceedings of the IEEE/CVF Conference on Computer Vision and Pattern Recognition (CVPR)},
  year      = {2024}
}

@inproceedings{karunratanakul2023gmd,
  title     = {Guided Motion Diffusion for Controllable Human Motion Synthesis},
  author    = {Karunratanakul, Korrawe and Preechakul, Konpat and Suwajanakorn, Supasorn and Tang, Siyu},
  booktitle = {Proceedings of the IEEE/CVF International Conference on Computer Vision (ICCV)},
  year      = {2023}
}

@inproceedings{xu2023interdiff,
  title     = {{InterDiff}: Generating {3D} Human-Object Interactions with Physics-Informed Diffusion},
  author    = {Xu, Sirui and Li, Zhengyuan and Wang, Yu-Xiong and Gui, Liang-Yan},
  booktitle = {Proceedings of the IEEE/CVF International Conference on Computer Vision (ICCV)},
  year      = {2023}
}

@article{peng2023hoidiff,
  title   = {{HOI-Diff}: Text-Driven Synthesis of {3D} Human-Object Interactions using Diffusion Models},
  author  = {Peng, Xiaogang and Xie, Yiming and Wu, Zizhao and Jampani, Varun and Sun, Deqing and Jiang, Huaizu},
  journal = {arXiv preprint arXiv:2312.06553},
  year    = {2023}
}

@article{romero2017mano,
  title   = {Embodied Hands: Modeling and Capturing Hands and Bodies Together},
  author  = {Romero, Javier and Tzionas, Dimitrios and Black, Michael J.},
  journal = {ACM Transactions on Graphics (SIGGRAPH Asia)},
  volume  = {36},
  number  = {6},
  year    = {2017}
}

@inproceedings{yang2022oakink,
  title     = {{OakInk}: A Large-Scale Knowledge Repository for Understanding Hand-Object Interaction},
  author    = {Yang, Lixin and Li, Kailin and Zhan, Xinyu and Wu, Fei and Xu, Anran and Liu, Liu and Lu, Cewu},
  booktitle = {Proceedings of the IEEE/CVF Conference on Computer Vision and Pattern Recognition (CVPR)},
  year      = {2022}
}

@inproceedings{liu2024geneoh,
  title     = {{GeneOH} Diffusion: Towards Generalizable Hand-Object Interaction Denoising via Denoising Diffusion},
  author    = {Liu, Xueyi and Yi, Li},
  booktitle = {International Conference on Learning Representations (ICLR)},
  year      = {2024}
}

@inproceedings{cha2024text2hoi,
  title     = {{Text2HOI}: Text-Guided {3D} Motion Generation for Hand-Object Interaction},
  author    = {Cha, Junuk and Kim, Jihyeon and Yoon, Jae Shin and Baek, Seungryul},
  booktitle = {Proceedings of the IEEE/CVF Conference on Computer Vision and Pattern Recognition (CVPR)},
  year      = {2024}
}

@inproceedings{christen2024diffh2o,
  title     = {{DiffH2O}: Diffusion-Based Synthesis of Hand-Object Interactions from Textual Descriptions},
  author    = {Christen, Sammy and Hampali, Shreyas and Sener, Fadime and Remelli, Edoardo and Hodan, Tomas and Sauser, Eric and Ma, Shugao and Tekin, Bugra},
  booktitle = {SIGGRAPH Asia Conference Papers},
  year      = {2024}
}

@inproceedings{li2025latenthoi,
  title={LatentHOI: On the Generalizable Hand Object Motion Generation with Latent Hand Diffusion.},
  author={Li, Muchen and Christen, Sammy and Wan, Chengde and Cai, Yujun and Liao, Renjie and Sigal, Leonid and Ma, Shugao},
  booktitle={Proceedings of the Computer Vision and Pattern Recognition Conference},
  pages={17416--17425},
  year={2025}
}

@article{zhang2026openhoi,
  title={Openhoi: Open-world hand-object interaction synthesis with multimodal large language model},
  author={Zhang, Zhenhao and Shi, Ye and Yang, Lingxiao and Ni, Suting and Ye, Qi and Wang, Jingya},
  journal={Advances in Neural Information Processing Systems},
  volume={38},
  pages={166582--166612},
  year={2026}
}

@inproceedings{xu2025intermimic,
  title={Intermimic: Towards universal whole-body control for physics-based human-object interactions},
  author={Xu, Sirui and Ling, Hung Yu and Wang, Yu-Xiong and Gui, Liang-Yan},
  booktitle={Proceedings of the Computer Vision and Pattern Recognition Conference},
  pages={12266--12277},
  year={2025}
}

@article{chen2024object,
  title={Object-centric dexterous manipulation from human motion data},
  author={Chen, Yuanpei and Wang, Chen and Yang, Yaodong and Liu, C Karen},
  journal={arXiv preprint arXiv:2411.04005},
  year={2024}
}

@misc{genesis2024,
  author  = {Genesis Authors},
  title   = {Genesis: A Universal and Generative Physics Engine for Robotics and Beyond},
  year    = {2024},
  month   = {December},
  url     = {https://github.com/Genesis-Embodied-AI/Genesis}
}

@misc{makoviichuk2021rlgames,
  author = {Denys Makoviichuk and Viktor Makoviychuk},
  title  = {rl-games: A High-Performance Framework for Reinforcement Learning},
  year   = {2021},
  month  = {May},
  url    = {https://github.com/Denys88/rl_games}
}

@article{schulman2017proximal,
  title={Proximal policy optimization algorithms},
  author={Schulman, John and Wolski, Filip and Dhariwal, Prafulla and Radford, Alec and Klimov, Oleg},
  journal={arXiv preprint arXiv:1707.06347},
  year={2017}
}

@article{dhariwal2021diffusion,
  title={Diffusion models beat gans on image synthesis},
  author={Dhariwal, Prafulla and Nichol, Alexander},
  journal={Advances in neural information processing systems},
  volume={34},
  pages={8780--8794},
  year={2021}
}

@article{madhoi,
  title   = {MAD-HOI: Masked Autoregressive Diffusion for Generating Articulated Hand Object Interactions from Text},
  author  = {Bal, Ananya and Sharma, Kartik and Lai, Ethan and Tiwari, Samyak and Dahiya, Liza and Chawla, Chaitanya and Jeni, Laszlo A},
  journal = {arXiv preprint arXiv:2608.10162},
  year    = {2026}
}

@article{loper2015smpl,
  title   = {{SMPL}: A Skinned Multi-Person Linear Model},
  author  = {Loper, Matthew and Mahmood, Naureen and Romero, Javier and Pons-Moll, Gerard and Black, Michael J.},
  journal = {ACM Transactions on Graphics},
  volume  = {34},
  number  = {6},
  pages   = {248:1--248:16},
  year    = {2015},
  publisher = {ACM}
}
